# Benchmark Generation Framework with Customizable Distortions for Image Classifier Robustness

Soumyendu Sarkar[†*]  Ashwin Ramesh Babu[†]  Sajad Mousavi[†]  Zachariah Carmichael[†]
Vineet Gundecha  Sahand Ghorbanpour  Ricardo Luna Gutierrez  Antonio Guillen
Avisek Naug
Hewlett Packard Enterprise, USA
{soumyendu.sarkar, ashwin.ramesh-babu, sajad.mousavi, zachariah.carmichael}@hpe.com
{vineet.gundecha, sahand.ghorbanpour, rluna, antonio.guillen, avisek.naug}@hpe.com

## Abstract

*We present a novel framework for generating adversarial benchmarks to evaluate the robustness of image classification models. Our framework allows users to customize the types of distortions to be optimally applied to images, which helps address the specific distortions relevant to their deployment. The benchmark can generate datasets at various distortion levels to assess the robustness of different image classifiers. Our results show that the adversarial samples generated by our framework with any of the image classification models, like ResNet-50, Inception-V3, and VGG-16, are effective and transferable to other models causing them to fail. These failures happen even when these models are adversarially retrained using state-of-the-art techniques, demonstrating the generalizability of our adversarial samples. We achieve competitive performance in terms of net $L_2$ distortion compared to state-of-the-art benchmark techniques on CIFAR-10 and ImageNet; however, we demonstrate our framework achieves such results with simple distortions like Gaussian noise without introducing unnatural artifacts or color bleeds. This is made possible by a model-based reinforcement learning (RL) agent and a technique that reduces a deep tree search of the image for model sensitivity to perturbations, to a one-level analysis and action. The flexibility of choosing distortions and setting classification probability thresholds for multiple classes makes our framework suitable for algorithmic audits.*

## 1. Introduction

Neural networks' susceptibility to adversarial perturbations has raised concerns about their reliability. Adversarial perturbations are slight alterations to input data that can cause neural networks to make confident yet incorrect predictions. Despite efforts to understand and counter adversarial perturbations, existing defense strategies have shown limited improvements in robust accuracy. This emphasizes the need for alternative approaches to evaluate and enhance neural network robustness. Recent research suggests that generating additional subsets from the main dataset through perturbations/augmentations can improve robustness in fully-supervised and semi-supervised settings [17]. To utilize the original training set more effectively, modifications are introduced. One popular recent approach, proposed by Hendrycks and Dietterich (2018), aims to evaluate model robustness and ultimately enhance it [17].

We propose a machine learning-driven adversarial data generator that introduces natural distortions to create an adversarial subset from an original dataset. Our approach formulates the generation of adversarial samples as a Markov Decision Process (MDP). By dividing the input sample into patches, we aim to identify and add distortions to the most vulnerable areas, leading to misclassification. Our generator utilizes an addition and removal mechanism, mimicking a deep tree search to find vulnerabilities and add noise in the right locations. Additionally, our method allows users to incorporate custom datasets and distortion types for generating adversarial samples.

As part of our work, we provide adversarial subsets derived from CIFAR-10 and ImageNet datasets. We evaluated the performance of adversarially trained models using state-of-the-art techniques from the literature on our dataset. The performance of these models on our dataset is noticeably lower than on the clean dataset and a competitor's benchmark [17]. We achieved an average $L_2$ value of 2.48 (evaluated over 1,000 ImageNet samples) and a maximum of 4.74. Our benchmark will assist future initiatives in building robust architectures, which is crucial considering the increasing concerns and requirements for robust deep-learning models.

---

[*]Corresponding author. [†]Equal contribution.

The main contributions of this paper are as follows:

- We propose a framework to generate adversarial benchmarks with a custom mix of distortions for evaluating the robustness of image classification models against both true negatives and false positives.
- We enable robustness audits for distortions characteristic of use cases at deployment for multiple distortion thresholds.
- We achieve competitive performance with the state-of-the-art on multiple metrics of minimum distortions needed for misclassification.
- We are competitive with the state-of-the-art on improving robustness with adversarial training.

## 2. Related Works

### 2.1. Data augmentation and adversarial samples for improving robustness

Several data augmentation techniques have been proposed to enhance the robustness of deep learning models. Cutout [9] masks out regions of input images which forces models to rely on alternative informative features. Mixup [54] generates virtual training samples by interpolating between pairs of images and labels, reducing overfitting and increasing robustness. Manifold Mixup [49] extends this idea by interpolating between feature representations. CutMix [53] combines Cutout and Mixup by replacing masked regions with patches from other images. AugMix [18] applies diverse augmentations to images, encouraging models to learn from a wide range of variations. Randaugment [7] applies random sequences of augmentation policies. RandConv [52] applies random convolutions as data augmentation. ALT [14] uses adversarially learned transformations to obtain both objectives of diversity and hardness at the same time. AutoAugment [6] and other recent works [33–35] uses Reinforcement Learning (RL) to discover optimal data augmentation policies. These techniques manipulate training data through various transformations, improving the models' robustness and generalization to adversarial perturbations. Besides RL has been used to solve various problems successfully [26, 36–42, 44].

### 2.2. Adversarial training for improved robustness

Recent research has explored various approaches to improve the robustness and out-of-distribution (OOD) performance of deep networks. Diffenderfer et al. [10] focused on compressing deep networks to enhance OOD robustness, demonstrating improved performance in handling OOD samples through network compression techniques. Kireev et al. [20] investigated the effectiveness of adversarial training against common corruption, identifying strengths and limitations of this approach. They explored the performance of adversarially trained models and suggested areas for improvement. Modas et al. [25] proposed PRIME, a framework that leverages primitive transformations during training to enhance robustness against common corruptions, achieving significant improvements in model performance on corrupted inputs. Wang et al. [50] introduced better diffusion models in adversarial training to enhance its effectiveness against adversarial attacks. Tian et al. [47] conducted a comprehensive analysis of the robustness of Vision Transformers (ViTs) towards common corruptions. Geirhos et al. [12] presented a study on the bias toward texture in ImageNet-trained Convolutional Neural Networks (CNNs), showing their reliance on texture rather than shape cues. Erichson et al. [11] developed NoisyMix, a framework that combines data augmentations, stability training, and noise injections to improve the robustness of deep neural networks.

### 2.3. Benchmark to evaluate robustness

Data augmentation techniques and benchmark datasets play a crucial role in evaluating and enhancing the robustness of image classification models. Hendrycks and Dietterich [17] introduced multiple datasets based on ImageNet and used as benchmarks for evaluating the robustness of models to input corruptions. **ImageNet-C** contains common visual corruptions applied to the ImageNet dataset and allows researchers to assess model performance under various types of visual distortions. **ImageNet-A** focuses on evaluating robustness to common image corruptions by providing a standardized evaluation environment. **ImageNet-P**, on the other hand, assesses the vulnerability of models to subtle perturbations by introducing imperceptible changes to deceive the models while maintaining visual similarity. The **Adversarial Robustness 101 (AR101)** benchmark [5] provides a comprehensive evaluation of model robustness against different attack types using the CIFAR-10 and CIFAR-100 datasets. PACS, Office-Home, MNIST-C, and WILDS benchmark datasets [1, 24, 27, 32] are designed to evaluate the domain adaptation and out-of-distribution robustness of the models. Lastly, the **Robustness via Dataset Manipulation (RoD)** [43] benchmark focuses on evaluating adversarial robustness against physical-world attacks by including real-world images with physical modifications. These benchmarks enable researchers to compare the performance of models and defense techniques in challenging scenarios.

## 3. Design of the Benchmark Generator

The evaluation for a machine learning model can be represented as $y = \text{argmax} f(x; \theta)$, where $x$ denotes the input image, $y$ represents the prediction, $\theta$ represents the model parameters, and the function $f$ represents the machine learning model's output,

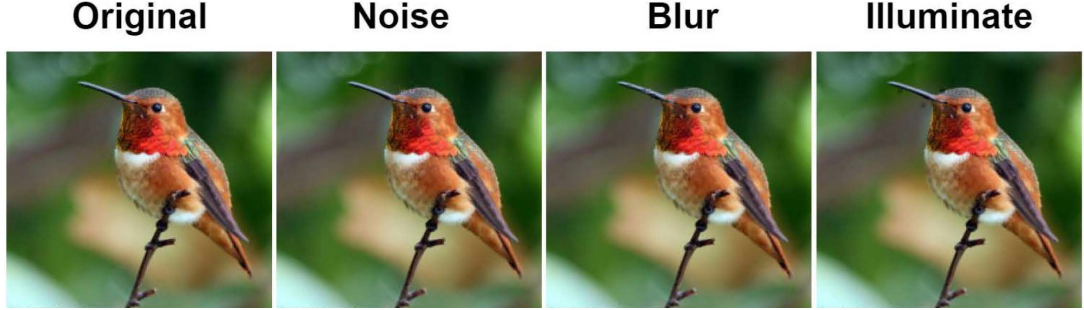

Figure 1. Adversarial samples with multiple distortion types (original picture from ImageNet)

## 3.1. Markov Decision Process (MDP) formulation

### 3.1.1 MDP for un-targeted attack

An un-targeted black-box adversarial sample generator, used for true negative evaluation, without access to the $\theta$, generates a perturbation $\delta$ such that $y_{true} \neq f(x+\delta; \theta)$. $L_p$ norms specify the distance between the original and the adversarial sample, $D(x, x+\delta)$. Our objective is to cause misclassification while keeping $D$ to a minimum.

State $S_t$ contains a number of lists related to the classification probability and sensitivity of the image regions. Action $A_t$ represent the perturbation to obtain the adversarial sample defined as:

$$A_t : x \to x + \delta_t, \qquad (1)$$

where $\delta_t$ defines the perturbation at time step $t$, or more specifically which patches of the original sample $x$ are going to be distorted. We define a probability dilution (PD) metric, which measures the extent to which the classification probability shifts from the ground truth to the other classes. The difference between the PD of the altered and the original image as a result of an action at each step ($\Delta$PD), is a measure of the effectiveness of the action. Moreover, the change in $L_2$ distance ($\Delta L_2$) as a measure of the distortion added is the cost for action. The reward is defined by the normalized PD as represented in equation 2.

$$R_t = \Delta PD_{norm} = -\Delta PD / \Delta L_2 \qquad (2)$$

The change in the distribution of the probabilities across classes is updated in the state vector at every step such that the RL agent can choose the optimum action at every step, maintaining the $L_p$ and the number of steps (queries).

### 3.1.2 MDP for targeted attack

A targeted black-box attack, used for false positive evaluation, without access to the $\theta$ generates a perturbation $\delta$ such that $y_{target} = f(x+\delta; \theta)$ s.t. $y_{target} \neq y_{true}$. $L_p$ norms specify the distance between the original and the adversarial sample, $D(x, x+\delta)$. Our objective is to cause misclassification while keeping $D$ to a minimum. The action $A_t$ with be defined as in equation 1.

We define a probability enhancement (PE) metric, which measures the extent to which the classification probability of the non-ground truth target class goes up. The difference between the PE of the altered image and the original image as a result of an action at each step ($\Delta$PE), is a measure of the effectiveness of the action. Moreover, the change in $L_2$ distance ($\Delta L_2$) as a measure of the distortion added is the cost for action. The reward is defined by the normalized PE as represented in equation 3.

$$R_t = \Delta PE_{norm} = \Delta PE / \Delta L_2 \qquad (3)$$

The change in the distribution of the probabilities across classes is updated in the state vector at every step such that the RL agent can choose the optimum action at every step, maintaining the $L_p$ and the number of step/queries.

## 3.2. Dual-action speedup for Deep Tree Search

### 3.2.1 Overview and Modification to MDP

In the proposed method, the input image is divided into square patches of size $n \times n$. For a true negative case, the sensitivity of the ground truth probability ($P_{GT}$) to addition and removal of distortion is computed for each patch. Based on this sensitivity information, our agent takes two actions at each step: select patches to which distortions are added and selected patches to which distortions are removed. In such a case we can define the state $S_t$ and action $A_t$ for timestep $t$ as:

$$S_t = S_t^+ + S_t^- \qquad (4)$$

$$A_t : x \to x + \delta_t^+ - \delta_t^-, \qquad (5)$$

where for timestep $t$, $S_t^+$ is the state after the add distortions perturbation $\delta_t^+$ is performed, and $S_t^-$ is the state after the remove distortions perturbation $\delta_t^-$ is applied.

This process is iteratively performed until the model misclassifies an image or until the budget for the number of maximum allowed steps is reached. In the case of mixed filter setting, the RL agent also needs to choose the optimal type of distortion filter for each step. For introducing the distortion at different threshold levels for untargeted adversarial samples, the process continues until the threshold level of distortion is reached.

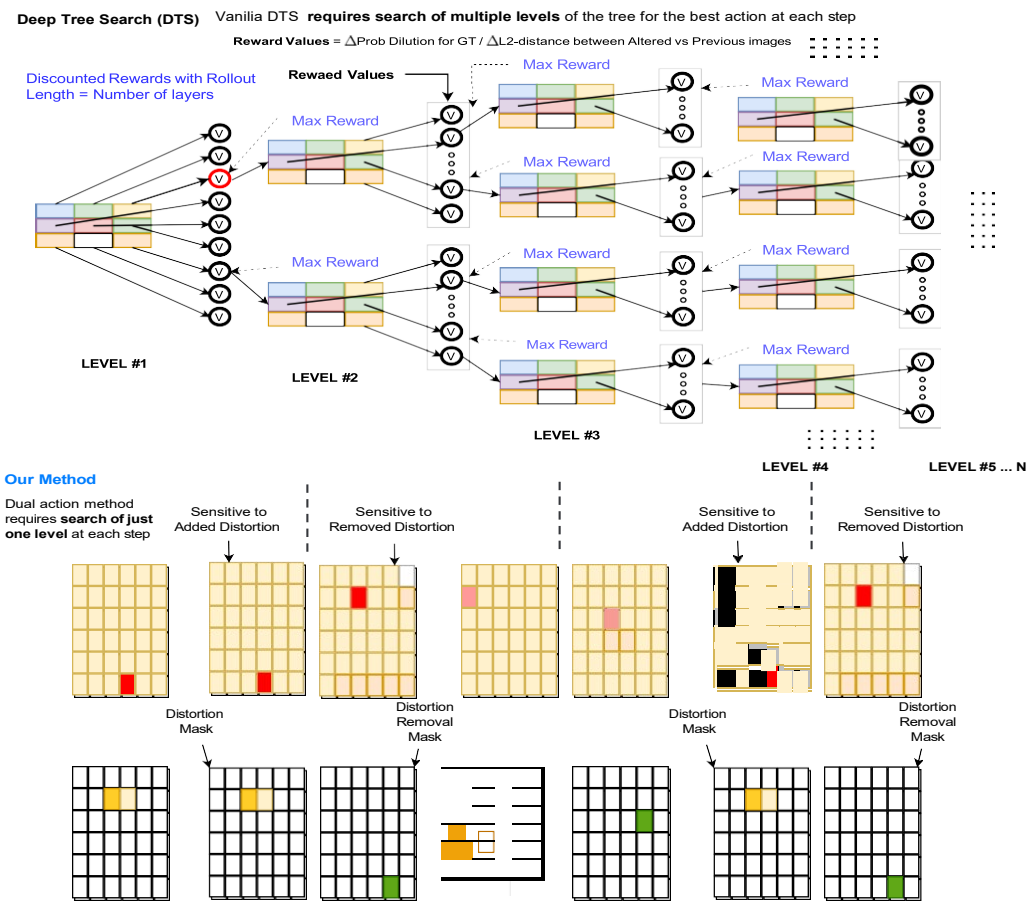

Figure 2. Dual-action architecture simplifying deep tree search

A similar technique is adopted for false positive benchmark generation with targeted adversarial samples, where the distortions are added to improve the classification probability of a non-ground truth class.

### 3.2.2 Intuition for dual-action

The idea of having two actions, addition and removal, is inspired by the limitations of the RL techniques used in board games. In that setting, the most effective moves are determined through a computationally expensive process called Deep Tree Search (DTS), which looks ahead multiple layers on a longer time horizon as the game progresses. However, unlike board games, in this problem, we have the ability to undo previous moves if we realize they were suboptimal. In the our framework, this is achieved by removing distortions added to patches in earlier steps and adding distortions to other patches, considering the current state of the modified image. This is similar to replaying all the moves in one step while analyzing the sensitivity of the image only at its current state, without performing a complete tree search.

By adopting this approach, we can significantly reduce the computational complexity from $O(N^d)$ to $O(N)$. Here, $N$ represents the computation complexity of evaluating one level and corresponds to the image size, while $d$ represents the depth of the tree search, which indicates how far ahead we look in the decision-making process.

### 3.2.3 Sensitivity Analysis

For the sensitivity analysis, distortion filters (masks) of size $n \times n$ are created with specific hyperparameters like distortion levels. These hyperparameters remain constant throughout the experiment. The filters are applied to square patches during training and validation to measure the change in the ground truth classification probability ($P_{GT}$). The hyperparameters of the distortion filters are chosen with minimal values to gradually introduce distortion and control the $L_p$ norm effectively. The distorted samples are con-strained to the range of $[0, 1]^d$, where $d$ is the dimensionality of the data. When multiple filters are available for the reinforcement learning (RL) agent to choose from, the hyperparameters are selected to have the same impact on the $L_p$ norm after applying any filter.

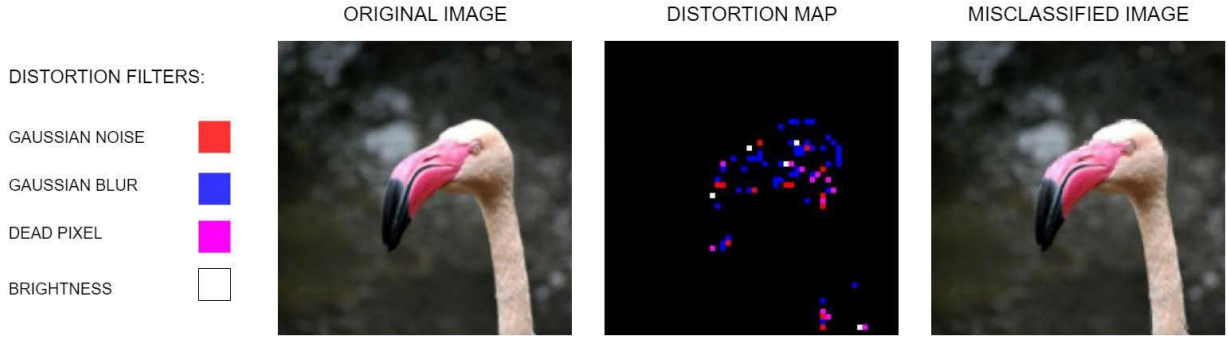

Figure 3. Mix of distortions for Adversarial Sample Generation

### 3.2.4 State Vector

The state vector was designed with the output of the image sensitivity analysis ordered based on the drift in $P_{GT}$ for patches during addition ($LIST^+$) and removal ($LIST^-$) of distortions. In addition, the classification probabilities of each class at every step ($LIST^p$) and the $L_p$ norm are included in the state vector.

## 3.3. Flexibility to use custom distortions

Our framework offers great versatility by allowing users to apply any type of distortion of their choice. The RL algorithm within the framework learns a policy that can adapt to different filters, ensuring that adversarial samples are generated with minimal distortion, denoted as $D$. Additionally, the algorithm can handle a combination of filters. At each step, the agent determines which filter (e.g., Gaussian noise, Gaussian blur, brightness adjustment) to use and the number of patches to which the filter should be applied. In our experiments, we explored multiple filters and presented four naturally occurring distortion filters in this paper. Figure 1 displays adversarial examples generated using different filters, while Figure 3 showcases adversarial examples generated with a mixture of various distortion filters.

## 4. Metrics and Experiments

We evaluate our proposed method with two different types of distortions: Gaussian noise and Gaussian blur. Since these types of common corruptions can be subtle or destructive, we generate data with five levels of severity $s$ and aggregate their scores. Clean error ($E^{clean}$) is defined as the top-1 misclassification of samples from the clean test set by evaluating the pre-existing classifier on the un-perturbed dataset. Corrupt error ($E^{corrupt}$) is defined as the top-1 misclassification of the samples from the corrupt dataset by evaluating the pre-existing classifier on the perturbed dataset. The performance of the classifier across the different severities level of corruption can be represented as:

$$CE^{corrupt} = \sum_{s=1}^{5} E_s^{corrupt} \quad (6)$$

$$\text{Accuracy}^{corrupt} = 1 - CE^{corrupt} \quad (7)$$

$$CE^{degradation} = \sum_{s=1}^{5} \left( E_s^{clean} - E_s^{corrupt} \right) \quad (8)$$

Furthermore, different corruptions pose different levels of difficulty as the effect of adding Gaussian noise, Gaussian blur, and illumination do not have the same impact on the sample. Note that in our results, for better robustness, we calculate the mean across the different corruption techniques used in this work (denoted as $m_{CE}$). Finally, accuracy degradation is the decline in the classifier performance when evaluated on the both clean and corrupted dataset.

Our benchmark is used to evaluate models from RobustBench [4], which is a reputable and continuously updated resource that both tracks and benchmarks adversarial robustness methods. The state-of-the-art models are selected by evaluating methods among thousands of papers on difficult benchmarks: $L_2$-constrained attacks, $L_\infty$-constrained attacks, and corruptions on standard image classification datasets. As RobustBench has built its reputation as a core scientific resource for tracking robustness progress, we treat the best-performing methods as the state-of-the-art in the literature. This is further substantiated as methods are included selectively: they cannot generally have non-zero gradients with respect to the input, have a fully deterministic forward pass, nor lack an optimization loop. It is known that the violation of these guidelines does not substantially improve robustness in general [2, 3].

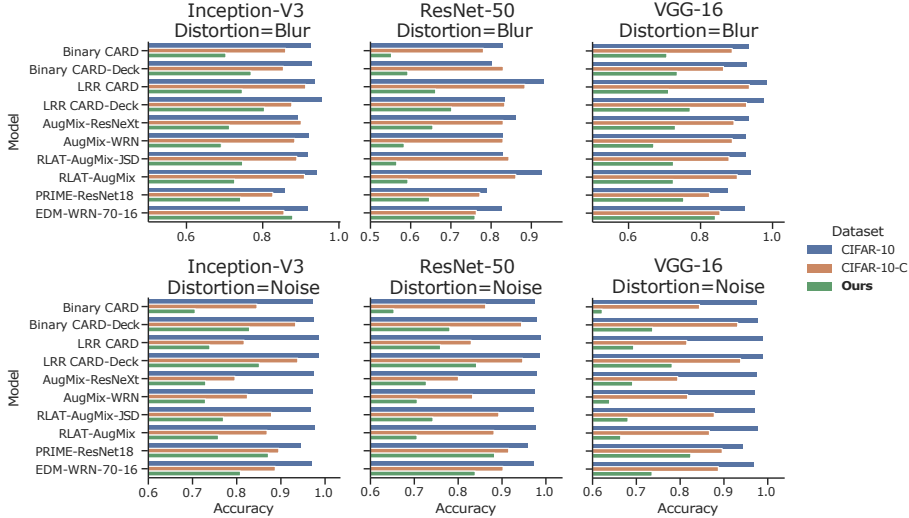

Figure 4. Evaluation of state-of-the-art robustness methods on corrupted versions of CIFAR-10: our corruptions with three victim models (ResNet-50, Inception-V3, and VGG-16) and CIFAR-10-C. Across two kinds of distortions, Gaussian noise and blur, our corrupted version of CIFAR-10 reduces accuracy more than CIFAR-10-C in most cases. Lower accuracy means better performance.

## 4.1. Compute Details

The computation for the complete pipeline is GPU-dependent and is efficiently batched and scaled on GPUs. Caching techniques were used for pre-computed information such as the noise masks for improved efficiency. Apollo servers with 8 V100 32GB GPUs were used for training and validation, as well as the evaluation of robustness methods. We processed 16 (images per GPU) × 8 (GPUs) = 128 images in a batch for the complete pipeline. See the supplemental material for additional details.

## 5. Results and Discussion

### 5.1. CIFAR-10

To validate the effectiveness of our generated benchmark, we compare the performance of state-of-the-art robustness methods between our distorted version of CIFAR-10 [23] and CIFAR-10-C [17]. CIFAR-10-C comprises distorted versions of the CIFAR-10 test set that are applied at five different severity levels. For a fair comparison, we compute the average $L_2$ distance between the original test set and the CIFAR-10-C test set for each type of distortion. We then employ our framework to generate distorted versions of those data splits for the approximate average $L_2$ of each CIFAR-10-C severity. Due to our sample generation procedure, we do not set a target $L_2$ (nor do the generators of CIFAR-10-C) so we must approximate the target average $L_2$. In experiments, we set generation parameters empirically and keep splits that have an average $L_2$ of within 25%. Often, especially with Gaussian blur, our average $L_2$ is far lower than that of CIFAR-10-C. See Appendix G for further details.

We select the top-10 ranked robustness methods, which includes state-of-the-art diffusion models, on CIFAR-10-C that are reported on the RobustBench benchmark [4] for evaluation: Binary CARD(-Deck) [10], LRR CARD(-Deck) [10], AugMix-ResNeXt [18], AugMix-WRN [18], RLAT-AugMix(-JSD) [20], PRIME-ResNet18 [25], and EDM-WRN-70-16 [50]. For each severity and victim model, we generate two sets of samples with Gaussian noise and Gaussian blur distortions, respectively. We consider VGG-16, Inception-V3, and ResNet-50 as the victim models in experiments. As discussed in Section 3, our framework does not generate a sample if the victim model misclassifies it initially. Hence, we generate distorted samples on a subset of the test set. For a fair comparison, we take the same subset from both CIFAR-10 and CIFAR-10-C to compute clean and corrupted performance, respectively. This sample-wise comparison ensures that harder samples are not excluded or easier samples are not included by one split or another. This is done by storing the indices of every sample in each split, including the original split, CIFAR-10-C split, and our split to prevent samples from inflating or deflating accuracy between splits. The results of these evaluations are shown in Figure 4. For each victim model and distortion, the scores on each CIFAR-10 test set are aggregated across all five levels of severity. For the blur distortion, we cause greater or equal degradation in performance than CIFAR-10-C across all robustness methods and victim models. The except lies with EDM-WRN-70-16 on samples generated with the Inception-V3 victim model, albeit marginally. Typically, the degradation value is much higher on ours and, sometimes, over double than that of CIFAR-10-C. For the noise distortion, we cause greater or equal degradation in performance than CIFAR-10-C across robustness methods and each victim model. We report the accuracy of the state-of-the-art robustness methods across five severity levels for the Gaussian noise and blur distortions in Appendix A – the accuracy degradation is proportional to the severity level.

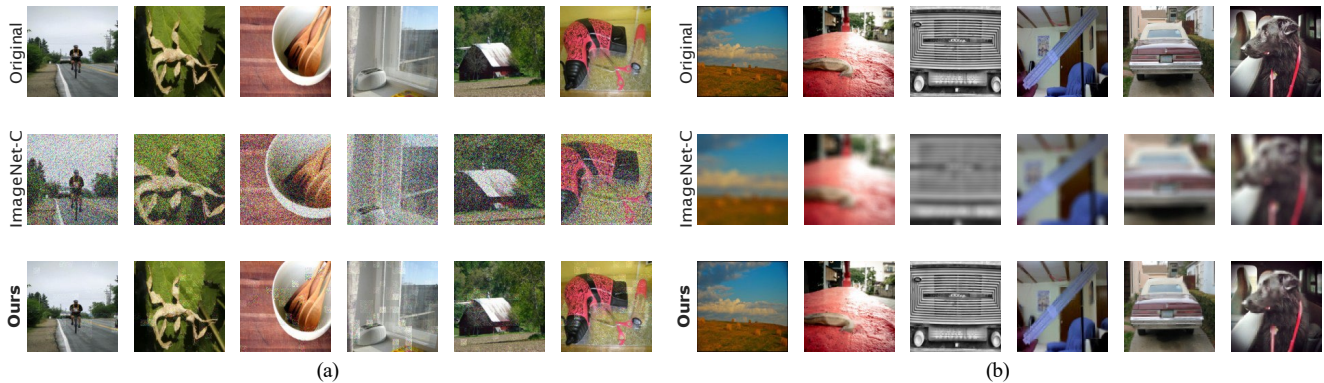

Figure 5. A subset of images from each of original ImageNet, ImageNet-C, and our distorted version of ImageNet. The images shown are for severity level 5 of the Gaussian (a) noise and (b) blur distortions. For the same severity level, images from ours retain much more clarity while being more challenging to classify.

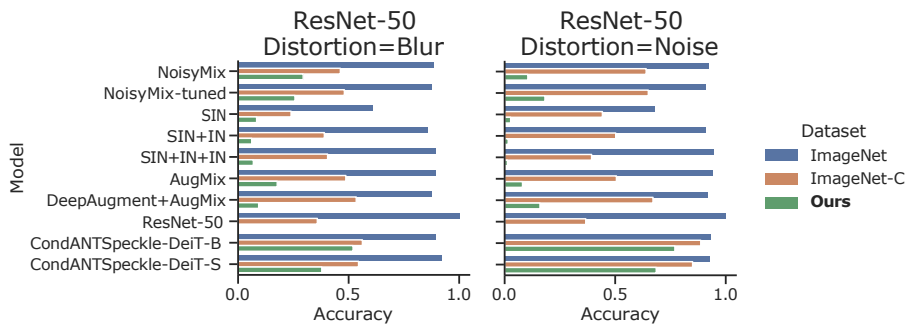

Figure 6. Evaluation of state-of-the-art robustness methods on corrupted versions of ImageNet: our corruptions and ImageNet-C. Our corrupted version of ImageNet reduces accuracy more than ImageNet-C in most cases. Lower accuracy means better performance.

## 5.2. ImageNet

To validate the effectiveness of our generated benchmark, we also compare the performance of state-of-the-art robustness methods between our distorted version of ImageNet and ImageNet-C. Figures 5a and 5b show some examples of the images in original ImageNet, ImageNet-C, and our distorted version of ImageNet. The images shown are for severity level 5 of the Gaussian noise and blur distortions, respectively. Note that ImageNet-C comes center-cropped and thus the full images are not shown. The evaluation here is conducted in the same manner as with CIFAR-10, ensuring that noise levels are similar and that a sample-wise comparison is conducted properly. We select the top-10 ranked robustness methods, which includes state-of-the-art ViTs, on ImageNet-C that are reported on the RobustBench benchmark for evaluation: DeepAugment+AugMix [16], CondANTSpeckle-DeiT-{S,B} [47], SIN(+IN(+IN)) [12], AugMix [18], standard ResNet-50, and NoisyMix(-tuned) [11].

The results of these evaluations are shown in Figure 6. Similar to our results on CIFAR-10, our distorted version of ImageNet results in greater accuracy degradation across the robustness methods than that of ImageNet-C. Notably, the mean $L_2$ level of ImageNet-C (99.3) is **69.0% higher than the mean $L_2$ level on our distorted version of ImageNet** (58.8) for Gaussian noise for the severity level of 5. Furthermore, the mean $L_2$ level of ImageNet-C (79.8) is *over 3× higher than the mean $L_2$ level on our distorted version of ImageNet* (25.6) for Gaussian blur. In both cases, we cause *greater accuracy degradation across all robustness models*. We report the accuracy of the state-of-the-art robustness method across five severity levels for the gaussian noise and blur distortions in Appendix A - the accuracy degradation is proportional to the severity level.

## 5.3. Results on Adversarial Retraining

Table 1 shows retrained robustness of the target model with our framework when compared to retraining with the other competitor approaches. The table presents the degradation error percentages for image classification architectures on the CIFAR-10-C dataset, comparing state-of-the-art techniques. The degradation errors for each technique are provided for three different models: ResNet-50, DenseNet, and Inception-V3.

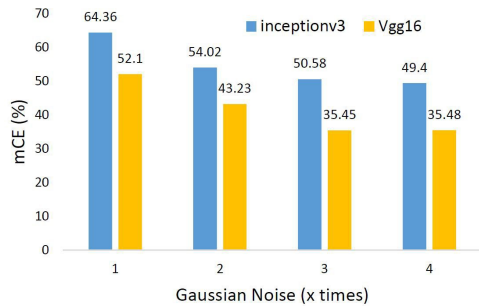

Figure 7. Evaluation of transferability of adversarial samples across other models

The results show that our framework outperforms the other techniques across all three models. For ResNet-50, we achieved a significantly lower degradation error of 6.0%, compared to Mixup (29.0%), CutMix (31.5%), and AugMix (13%). Similarly, for DenseNet and Inception-V3, our framework also demonstrates superior performance, with degradation errors of 11% and 9.5%, respectively, compared to the other techniques. These findings suggest that our framework effectively has the lowest degradation errors in image classification tasks on the CIFAR-10-C dataset, surpassing the performance of other state-of-the-art techniques like Mixup, CutMix, and AugMix.

Table 1. Degradation error % for image classification architectures on CIFAR-10-C for state-of-the-art techniques. For fairness, all of the techniques were evaluated with the same seed.

| Model | Mixup | CutMix | AugMix | Ours |
|---|---|---|---|---|
| ResNet-50 | 29.0 | 31.5 | 13 | **6.0** |
| DenseNet | 24.0 | 33.5 | 15 | **11** |
| Inception-V3 | 29 | 23 | 11.5 | **9.5** |
| Mean | 27.3 | 29.3 | 13.1 | 8.83 |

## 5.4. Transferability across different models

Table 2 represents the ability to transfer the adversarial samples across other primitive models. The adversarial samples are generated to deceive the pre-trained model shown in each row and are tested on the model shown in each column.

From the table, it can be understood that adversarial samples that were generated and evaluated on the same models have 0 accuracy. Furthermore, these adversarial samples still have a significant impact on the other primitive models showing the ability of the proposed method to generalize well. The values are averaged across both Gaussian blur and Gaussian noise types of distortions. Figure 7 illustrates the transferability of samples generated using the ResNet-50 model with the ImageNet dataset. These samples were tested on Inception-V3 and Vgg16 models under various noise levels. It can be observed that the samples generated by ResNet-50 still exhibit substantial correlation errors across different models. Also, as the noise level increases, the performance tends to decrease.

Table 2. Transferability of adversarial samples generated from CIFAR-10 across other primitive models. The values represent the classification accuracy mCE.

| | | ResNet-50 | Inception-V3 | VGG-16 |
|---|---|---|---|---|
| Victim | ResNet-50 | 0 | 12.19 | 8.93 |
| | Inception-V3 | 20.17 | 0 | 12.16 |
| | VGG-16 | 16.90 | 16.70 | 0 |

## 6. Limitations

The proposed method focuses on vulnerabilities of image classifiers from distortions present at deployment by providing the customization option. Our results with CIFAR-10-C benchmark show that our method is more effective in identifying vulnerabilities with optimal distortions which are generalizable across models. The nature of the distortion filters used by our model uncovers the broad vulnerabilities of the deployed model, but does not enable unnatural artifacts.

## 7. Conclusion and Future Work

This paper presents a novel approach to address the challenge of evaluating and improving the robustness of neural networks against adversarial perturbations. The proposed ML-driven adversarial data generator introduces naturally occurring distortions to the original dataset, creating an adversarial subset. By formulating the problem as an MDP, the generator effectively identifies and adds distortions to the most vulnerable areas of the input. This approach demonstrates competitive performance with state-of-the-art techniques, providing a benchmark for evaluating the robustness of image classification models. Additionally, the framework allows for the inclusion of custom distortion types, adversarial thresholds, and datasets, enabling tailored evaluations and audits for specific use cases. The results highlight the importance of building robust deep-learning models and offer valuable insights for future research and development in this area. Overall, this work contributes to the advancement of reliable and resilient deep learning architectures through the generation of adversarial benchmarks and the exploration of improved adversarial training methods. In the future, we will include evaluations on additional naturally occurring perturbations.

# Supplemental Section

## A. Additional ImageNet Results

We report the accuracy of state-of-the-art **Robust-Bench** [4] robustness methods across five severity levels for the Gaussian noise and blur distortions in Figure 8. The robustness methods perform worse on our split. The accuracy degradation is proportional to the severity level. Notably, the mean $L_2$ level of ImageNet-C (99.3) is **69.0% higher than the mean $L_2$ level on our distorted version of ImageNet** (58.8) for Gaussian noise for the intensity level of 5. Furthermore, the mean $L_2$ level of ImageNet-C (79.8) is **over 3× higher than the mean $L_2$ level on our distorted version of ImageNet** (25.6) for Gaussian blur. In both cases, we cause **greater accuracy degradation across all robustness models**.

### A.1. Evaluating on different filters

| Patch Size | AVG.Q | Max.$L_2$ (avg.) | ASR % |
|---|---|---|---|
| **Gaussian Noise** | **166** | **4.74(2.48)** | 100 |
| Illumination | 266 | 6.94(3.93) | 100 |
| Gaussian Blur | 201 | 9.3(5.33) | 100 |
| Dead Pixel | 75 | 21.16(13.58) | 100 |

Table 3. Comparison of maximum L2, average queries and Average Success rate with different **distortion filters with RLAB**. (avg) represents average L2 over all samples **Dataset:** ImageNet, **Model:** ResNet-50

Table 3 represents the performance of the proposed approach with different filters. From the table, it can be observed that the approach generates the best results for Gaussian noise followed by the illumination filter for the ImageNet dataset. For brightness, the best results were obtained for the brightness value of -0.1 with values ranging between (-1, 1). For DeadPixel, the percentage of pixels to be dropped for a given patch was set to 50 percent. For Gaussian blur, the standard deviation was set to 1. The standard deviation controls the amount of blurring with a larger value (> 1) creating significantly higher blurring compared to a smaller value. The performance of the Gaussian noise better than the other filters could be due to the strong nature of noise-based distortion proved in numerous works [28, 31].

## B. Additional CIFAR-10 Results

We report the accuracy of state-of-the-art **Robust-Bench** [4] robustness methods across all severity levels for the Gaussian noise distortion in Figure 9. Across the Inception-V3 and ResNet-50 victim models, robustness methods perform worse on our split on average. The accuracy degradation is proportional to the severity level. As explained in the main text, the samples generated with the VGG-16 victim model did not transfer for the Gaussian noise distortion.

## C. Error Bars

Here, we reported error bars for applicable experiments. For the robustness experiments comparing with CIFAR-10-C and ImageNet-C, the robustness models from Robust-Bench have a single set of weights provided, evaluations on CIFAR-10-C and ImageNet-C are deterministic (the splits are fixed), and evaluations on our distorted versions of CIFAR-10 and ImageNet are deterministic (the splits are fixed). Thus, *there are no error bars to report*. For the retraining,

## D. RobustBench and the State-of-the-Art in Adversarial Robustness

**RobustBench** [4] is a reputable and continuously updated resource that both tracks and benchmarks adversarial robustness methods. The state-of-the-art is selected by evaluating methods among thousands of papers on difficult benchmarks: $L_2$-constrained attacks, $L_\infty$-constrained attacks, and corruptions on standard image classification datasets. As RobustBench has built its reputation as the core scientific resource for tracking robustness progress, we treat the best-performing methods as the state-of-the-art in the literature. This is further substantiated as methods are included selectively: they cannot generally have non-zero gradients with respect to the input, have a fully deterministic forward pass, nor lack an optimization loop. It is known that the violation of these guidelines does not substantially improve robustness in general [2,3]. In our experiments, our distorted version of ImageNet and CIFAR-10 cause further degradation on these state-of-the-art methods than the challenging benchmarks evaluated within RobustBench. This degradation also is realized with less distortion (lower $L_2$) than these standard benchmarks.

## E. Benchmark Generation Algorithm

**Initialization:** Policy parameters
**Input:** Validation set, number of iterations $Max_{iter}$ = 3500
**Output:** Optimized policy for Dueling DQN

image **in** validation set Load the image  Calculate reward $R_t$ and advantage $\hat{A}_t$ based on current value function  Calculate sensitivity of ground truth classification probability $P_{GT}$ to change in

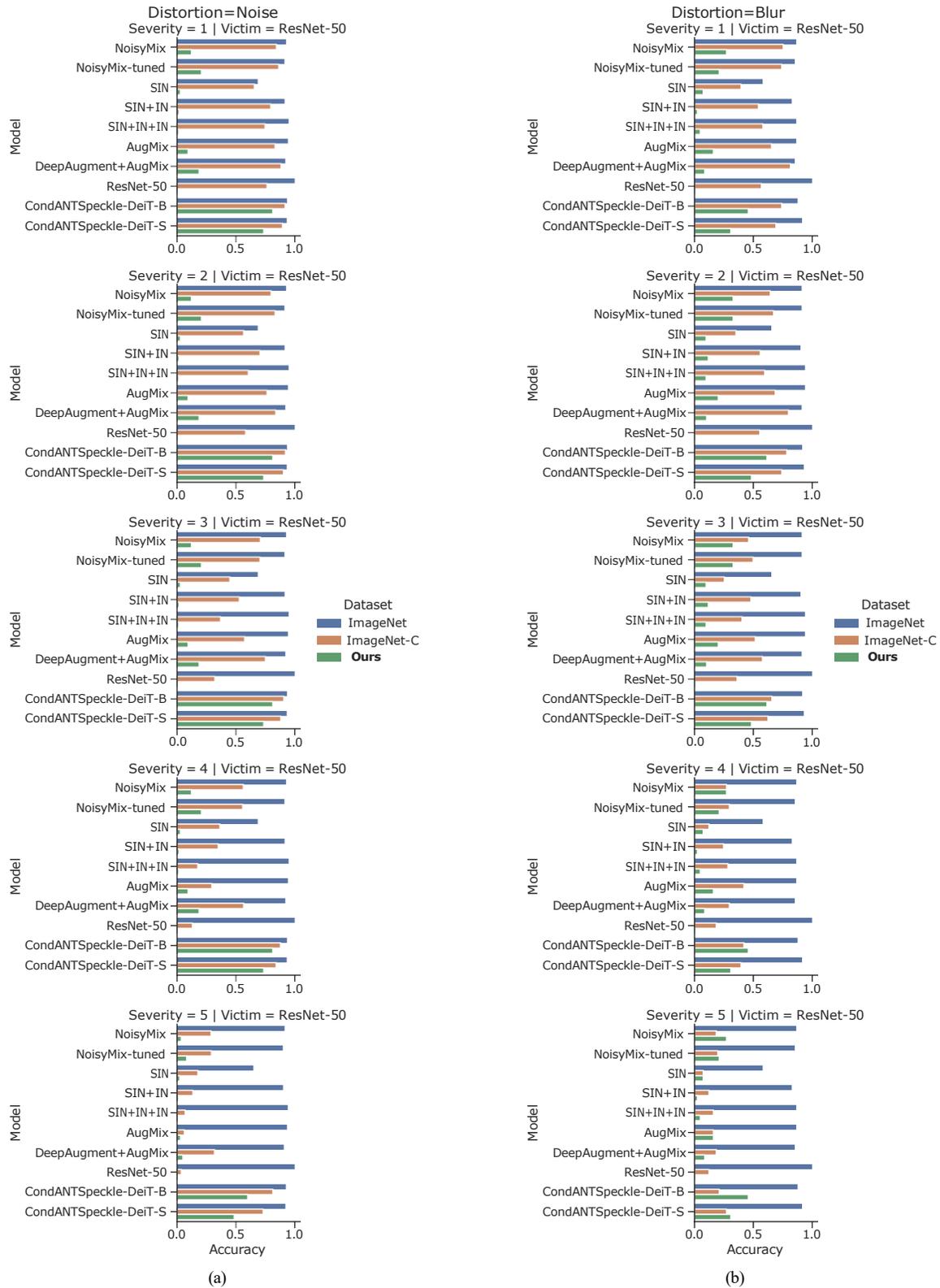

Figure 8. Robustness method scores on ImageNet, ImageNet-C, and our distorted version of ImageNet across all severity levels for the (a) Gaussian noise and (b) blur distortions. The robustness methods perform worse on our split.

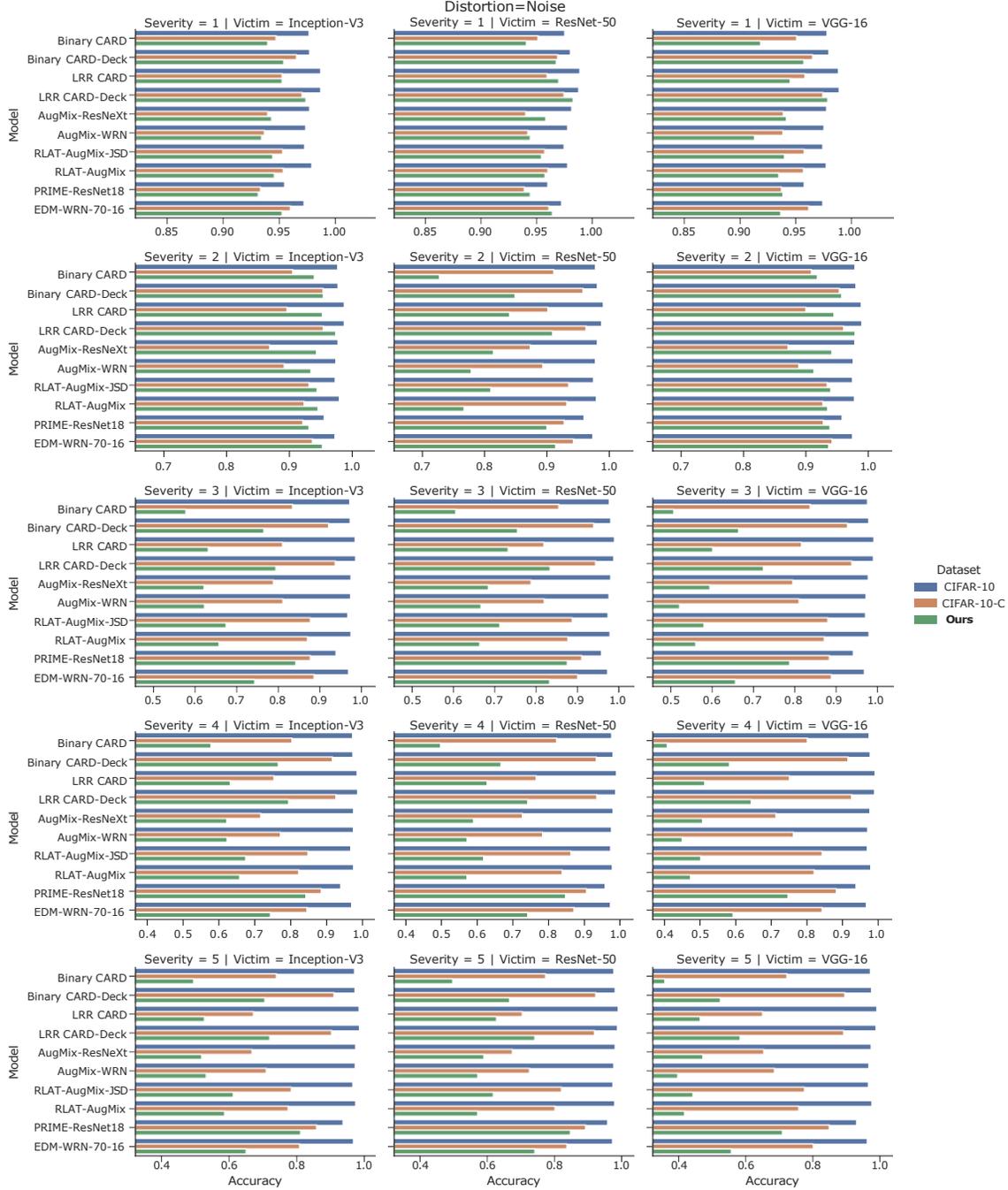

Figure 9. Robustness method scores on CIFAR-10, CIFAR-10-C, and our distorted version of CIFAR-10 across all severity levels for the Gaussian noise distortion. Across the Inception-V3 and ResNet-50 victim models, robustness methods perform worse on our split on average.

distortion for square patches $i \leftarrow 0$ $Pred_{fstep} \leftarrow 1 - P_{GT}$ $Pred_{GT} == Pred_{fstep}$ **and** $i < Max_{iter}$ Collect the set of trajectories (state, action) by running policy $\pi_k = \pi(\theta_k)$ in the environment $\rightarrow$ action $(N_{add\_dist}, N_{rem\_dist})$ Generate adversarial image $x'$ with respect to the distortion type and the chosen patch (i.e., select the action and take the action) Calculate reward $R_t$ and TD error Update the the RL network policy Perform prediction on $x' \rightarrow Pred_{fstep}$ $i \leftarrow i + 1$

Algorithm E shows the pipelines for benchmark generation.

## F. Dataset Information

**ImageNet** The ImageNet dataset [8] is provided contingent on terms that are specified on the ImageNet website. It is known that ImageNet contains web-scraped images containing people without their consent, which can be considered to contain personally identifiable information.

**ImageNet-C** The ImageNet-C dataset [17] has a Creative Commons Attribution 4.0 International license. It is known that ImageNet (and in turn its derivatives) contains web-scraped images containing people without their consent, which can be considered to contain personally identifiable information.

**CIFAR-10** The CIFAR-10 dataset [22] is provided without a license.

**CIFAR-10-C** The CIFAR-10-C dataset [17] has a Creative Commons Attribution 4.0 International license.

## G. Accessibility, Documentation, and License

Our dataset documentation can be found at `https://<redacted>.github.io/<redacted>` and code at `https://github.com/<redacted>`. The documentation includes how to load the generated data. We use Zenodo to associate a DOI with the repository. There is schema.org structured metadata provided on the website as well. The benchmark is made available under the MIT license, which is included with the GitHub repository.

We plan to continue hosting the benchmark through GitHub as-is and provide maintenance as necessary. We the authors bear all responsibility in case of violation of rights, etc.

## H. Experimental Details

**ImageNet** Experiments on ImageNet use its validation set of 50,000 samples. For robustness experiments involving RobustBench, the only preprocessing done is to scale the pixel values by 255 so that values lie between 0 and 1, as is standard for the pretrained models in RobustBench [4]. For other experiments, z-score normalization is applied with statistics taken from the train split.

**ImageNet-C** Experiments on ImageNet-C use its validation set of 10,000 samples. The ImageNet-C distortions do not include Gaussian blur, however, it does contain various types of blur. In experiments, we compare results on Gaussian blur distortion in our framework to defocus blur in ImageNet-C, which are more similar than the motion, glass, or zoom blurs. For robustness experiments involving RobustBench, the only preprocessing done is to scale the pixel values by 255 so that values lie between 0 and 1, as is standard for the pretrained models in RobustBench [4]. For other experiments, z-score normalization is applied with statistics taken from the train split. In RobustBench, there is a subset of the validation splits comprising 5,000 samples that is used for evaluation, rather than all 50,000.

**ImageNet-Ours** ImageNet-Ours follows the same train, validation, and test splits as the original ImageNet dataset [8]. The distortions in the dataset are generated on the validation split of ImageNet.

We provide this dataset on Zenodo: `https://zenodo.org/<redacted>`

For all our experiments, we used a patch size of 8×8 and the patches were non-overlapping to reduce the action space. We observed that increasing the size of the patch eventually increased the $L_2$ while it decreased the total number of steps to generate the adversarial sample. As we had a bound on the $L_2$, using the patch size of 8×8 gave us favorable results. The noise intensity level 1 is assumed to be the adversarial sample at the time of misclassification. We further increase the noise level by 2×, 3×, 4×, and 5× the noise at level 1 for the experiments.

**CIFAR-10** We use the standard CIFAR-10 train and test splits [22]. Experiments on CIFAR-10 use its test set of 10,000 samples. For robustness experiments involving RobustBench, the only preprocessing done is to scale the pixel values by 255 so that values lie between 0 and 1, as is standard for the pretrained models in RobustBench [4]. For other experiments, z-score normalization is applied with statistics taken from the train split.

**CIFAR-10-C** CIFAR-10-C follows the same train and test splits as the original CIFAR-10 dataset [22]. Experiments on CIFAR-10-C use its distorted versions of the CIFAR-10 test set, each of which is 10,000 samples. There is one such test set for five distortion levels and various types of distortions (corruptions), which contains the same corruptions that we apply in our framework – the results on these splits can be compared directly. For robustness experiments involving RobustBench, the only preprocessing done is to scale the pixel values by 255 so that values lie between 0 and 1, as is standard for the pretrained models in RobustBench [4]. For other experiments, z-score normalization is applied with statistics taken from the train split.

**CIFAR-10-Ours** CIFAR-10-Ours follows the same train and test splits as the original CIFAR-10 dataset [22]. The distortions in the dataset are generated on the test split of CIFAR-10.

We provide this dataset on Zenodo: `https://zenodo.org/<redacted>`.

For all our experiments, we used a patch size of 2×2 and the patches were non-overlapping to reduce the action space. We observed that increasing the size of the patch eventually increased the $L_2$ while it decreased the total number of steps to generate the adversarial sample. As we had a bound on the $L_2$, using the smallest patch size of 2 ×2 gave us favorable results. The noise intensity level 1 is assumed to be the adversarial sample at the time of misclassification. We further increase the noise level by 2×, 3×, 4×, and 5× the noise at level 1 for the experiments.

**Comparing Our Distorted Datasets with CIFAR-10-C/ImageNet-C** For a fair comparison, we compute the average $L_2$ distance (listed in main text) between the original test/validation sets and the CIFAR-10-C/ImageNet-C test/validation sets for each type of distortion. We then employ our framework to generate distorted versions of those data splits for the approximate average $L_2$ of each CIFAR-10-C/ImageNet-C severity. Due to our sample generation procedure, we do not set a target $L_2$ (nor do the generators of CIFAR-10-C/ImageNet-C) so we must approximate the target average $L_2$. In experiments, we set generation parameters empirically and keep splits that have an average $L_2$ of within 25%. Often, especially with Gaussian blur, our average $L_2$ is far lower than that of CIFAR-10-C/ImageNet-C. For instance, the mean $L_2$ level of ImageNet-C (99.3) is 69.0% higher than the mean $L_2$ level on our distorted version of ImageNet (58.8) for Gaussian noise, yet we cause greater accuracy degradation for robustness models.

We also perform a fair sample wise comparison such that harder samples are not excluded or easier samples are not included by one split or another. This is done by storing the indices of every sample in each split, including the original split, CIFAR-10-C/ImageNet-C split, and our split. The intersection between these three sets of indices is used for evaluation. To reiterate, this prevents sample difficulty from inflating or deflating accuracy between splits.

**Standard Models** We evaluated the performance of standard models such as VGG-16 [45], Inception-v3, and ResNet-50 [15, 46] and evaluated them against data generated with our adversarial generator. Table 2 in the main text shows the evaluation results across standard models as victims.

**Robustness Models** All robustness methods are available through the **RobustBench** [4] library. This includes their software licenses, neural network parameters, and neural network architectures implemented in `PyTorch` [30]. To select which robustness models to use, we evaluated every single available model (all threat models) on both CIFAR-10-C and ImageNet-C – this was done as not all models were evaluated on all RobustBench benchmarks. The top-10, as evaluated by the mean accuracy across all five severities on each of CIFAR-10-C and ImageNet-C, were selected to evaluate on our generated benchmarks. Citations for each selected method are present in the main text. We consider these to be the state-of-the-art as RobustBench is reputable and continuously updated.

### I. Compute Resources

Experiments were conducted on a single machine within an internal cluster. The machine contains eight NVIDIA Tesla V100 32GB GPUs, 756GB of DDR4 2933 MT/s RAM, and an Intel(R) Xeon(R) Gold 6248 CPU @ 2.50GHz (2 sockets, 2 cores/socket, and 2 threads/core for a total of 80 logical cores).

### J. Software Libraries

This work was made possible with the `RobustBench` [4] (MIT License), `PyTorch` [30] (BSD-Style), `seaborn` [51] (BSD 3-Clause), `pandas` [29] (BSD 3-Clause License), `matplotlib` [19] (Various Licenses), and `jupyter` [21] (BSD 3-Clause License) libraries for `Python` [48] (PSF License).

### K. Other Reproducibility Details

The code is seeded for reproducibility and contains every seed used. For experiments involving RobustBench, all model architectures and weights are available for download through the `RobustBench` [4] library. The remaining reproducibility information is detailed in the other sections of this document.

### L. Complete Benchmark Results

We include tables of every single model evaluated in our study for both CIFAR-10 and ImageNet. Please see the tables following the references.

| Distortion | Victim Model | Dataset Model ID | **Ours** | ImageNet | ImageNet-C |
|---|---|---|---|---|---|
| Blur | ResNet-50 | AlexNet | 0.2232 | 0.5146 | 0.1280 |
| | | Erichson2022NoisyMix | 0.4307 | 0.8750 | 0.4825 |
| | | Erichson2022NoisyMix_new | 0.4018 | 0.8946 | 0.4982 |
| | | Geirhos2018_SIN | 0.2146 | 0.6199 | 0.2512 |
| | | Geirhos2018_SIN_IN | 0.2583 | 0.8542 | 0.4192 |
| | | Geirhos2018_SIN_IN_IN | 0.2699 | 0.8794 | 0.4091 |
| | | Hendrycks2020AugMix | 0.3654 | 0.8870 | 0.5162 |
| | | Hendrycks2020Many | 0.2894 | 0.8719 | 0.5486 |
| | | Salman2020Do_50_2_Linf | 0.3801 | 0.6867 | 0.2349 |
| | | Standard_R50 | 0.0000 | 1.0000 | 0.3794 |
| | | Tian2022Deeper_DeiT-B | 0.5966 | 0.8801 | 0.5735 |
| | | Tian2022Deeper_DeiT-S | 0.5191 | 0.9105 | 0.5613 |

Table 4. All ImageNet accuracy scores for the Blur distortion and the victim model ResNet-50.

| Distortion | Victim Model | Dataset Model ID | **Ours** | ImageNet | ImageNet-C |
|---|---|---|---|---|---|
| Noise | ResNet-50 | AlexNet | 0.1932 | 0.6756 | 0.1254 |
| | | Erichson2022NoisyMix | 0.3595 | 0.9251 | 0.6477 |
| | | Erichson2022NoisyMix_new | 0.4180 | 0.9107 | 0.6566 |
| | | Geirhos2018_SIN | 0.1662 | 0.6850 | 0.4467 |
| | | Geirhos2018_SIN_IN | 0.2321 | 0.9136 | 0.5038 |
| | | Geirhos2018_SIN_IN_IN | 0.2277 | 0.9482 | 0.3928 |
| | | Hendrycks2020AugMix | 0.3277 | 0.9424 | 0.5075 |
| | | Hendrycks2020Many | 0.3697 | 0.9194 | 0.6786 |
| | | Salman2020Do_50_2_Linf | 0.2586 | 0.7964 | 0.3117 |
| | | Standard_R50 | 0.0000 | 1.0000 | 0.3660 |
| | | Tian2022Deeper_DeiT-B | 0.8167 | 0.9338 | 0.8903 |
| | | Tian2022Deeper_DeiT-S | 0.7553 | 0.9309 | 0.8555 |

Table 5. All ImageNet accuracy scores for the Noise distortion and the victim model ResNet-50.

| Distortion | Victim Model | Dataset Model ID | **Ours** | CIFAR-10 | CIFAR-10-C |
|---|---|---|---|---|---|
| Blur | Inception-V3 | Addepalli2021Towards_RN18 | 0.5638 | 0.6333 | 0.5618 |
| | | AGAT [13] | 0.2723 | 0.7382 | 0.6408 |
| | | Addepalli2021Towards_WRN34 | 0.5515 | 0.5989 | 0.5618 |
| | | Addepalli2022Efficient_RN18 | 0.5814 | 0.6279 | 0.5906 |
| | | Addepalli2022Efficient_WRN_34_10 | 0.6442 | 0.6921 | 0.6228 |
| | | Andriushchenko2020Understanding | 0.4899 | 0.5467 | 0.4871 |
| | | Augustin2020Adversarial | 0.6963 | 0.7941 | 0.7257 |
| | | Augustin2020Adversarial_34_10 | 0.7234 | 0.8170 | 0.7102 |
| | | Augustin2020Adversarial_34_10_extra | 0.7268 | 0.8571 | 0.7904 |
| | | Carmon2019Unlabeled | 0.5893 | 0.7265 | 0.6288 |
| | | Chen2020Adversarial | 0.6118 | 0.5943 | 0.5858 |
| | | Chen2020Efficient | 0.5519 | 0.6257 | 0.5834 |
| | | Chen2021LTD_WRN34_10 | 0.5804 | 0.6309 | 0.5665 |
| | | Chen2021LTD_WRN34_20 | 0.6172 | 0.6715 | 0.5929 |
| | | Cui2020Learnable_34_10 | 0.6444 | 0.6688 | 0.6388 |
| | | Cui2020Learnable_34_20 | 0.6051 | 0.6367 | 0.6100 |
| | | Dai2021Parameterizing | 0.5960 | 0.6838 | 0.5880 |
| | | Debenedetti2022Light_XCiT-L12 | 0.6667 | 0.7148 | 0.6699 |
| | | Debenedetti2022Light_XCiT-M12 | 0.6549 | 0.7211 | 0.6599 |
| | | Debenedetti2022Light_XCiT-S12 | 0.6458 | 0.6960 | 0.6301 |
| | | Diffenderfer2021Winning_Binary | 0.5088 | 0.8770 | 0.8185 |
| | | Diffenderfer2021Winning_Binary_CARD_Deck | 0.6158 | 0.8933 | 0.8129 |
| | | Diffenderfer2021Winning_LRR | 0.5735 | 0.8830 | 0.8756 |
| | | Diffenderfer2021Winning_LRR_CARD_Deck | 0.6538 | 0.9263 | 0.8484 |
| | | Ding2020MMA | 0.5873 | 0.6492 | 0.5769 |
| | | Engstrom2019Robustness | 0.6995 | 0.8019 | 0.6855 |
| | | Gowal2020Uncovering | 0.6728 | 0.7533 | 0.6838 |
| | | Gowal2020Uncovering_28_10_extra | 0.6213 | 0.7092 | 0.6246 |
| | | Gowal2020Uncovering_34_20 | 0.5401 | 0.6164 | 0.5698 |
| | | Gowal2020Uncovering_70_16 | 0.5753 | 0.6429 | 0.5850 |
| | | Gowal2020Uncovering_70_16_extra | 0.6293 | 0.7378 | 0.6251 |
| | | Gowal2020Uncovering_extra | 0.7661 | 0.8930 | 0.7805 |
| | | Gowal2021Improving_28_10_ddpm_100m | 0.6565 | 0.7039 | 0.6406 |
| | | Gowal2021Improving_70_16_ddpm_100m | 0.6601 | 0.7105 | 0.6553 |
| | | Gowal2021Improving_R18_ddpm_100m | 0.6300 | 0.6743 | 0.6231 |
| | | Hendrycks2019Using | 0.5486 | 0.6242 | 0.5549 |
| | | Hendrycks2020AugMix_ResNeXt | 0.5853 | 0.8259 | 0.8462 |
| | | Hendrycks2020AugMix_WRN | 0.5450 | 0.8728 | 0.8207 |
| | | Huang2020Self | 0.5707 | 0.6407 | 0.5783 |
| | | Huang2021Exploring | 0.6460 | 0.7403 | 0.6818 |
| | | Huang2021Exploring_ema | 0.6855 | 0.8141 | 0.6951 |
| | | Huang2022Revisiting_WRN-A4 | 0.6667 | 0.8182 | 0.6964 |
| | | Jia2022LAS-AT_34_10 | 0.6181 | 0.6649 | 0.5943 |
| | | Jia2022LAS-AT_70_16 | 0.6218 | 0.6764 | 0.6152 |
| | | Kang2021Stable | 0.7161 | 0.8240 | 0.7176 |
| | | Kireev2021Effectiveness_AugMixNoJSD | 0.5177 | 0.8734 | 0.8172 |
| | | Kireev2021Effectiveness_Gauss50percent | 0.6350 | 0.8585 | 0.7602 |
| | | Kireev2021Effectiveness_RLAT | 0.6537 | 0.8053 | 0.7099 |
| | | Kireev2021Effectiveness_RLATAugMix | 0.6319 | 0.8608 | 0.8241 |
| | | Kireev2021Effectiveness_RLATAugMixNoJSD | 0.6034 | 0.8983 | 0.8522 |
| | | Modas2021PRIMEResNet18 | 0.6212 | 0.7549 | 0.7434 |
| | | Pang2020Boosting | 0.5764 | 0.6558 | 0.5946 |
| | | Pang2022Robustness_WRN28_10 | 0.6207 | 0.6792 | 0.5975 |
| | | Pang2022Robustness_WRN70_16 | 0.5985 | 0.6586 | 0.6046 |
| | | Rade2021Helper_R18_ddpm | 0.6436 | 0.7489 | 0.6396 |
| | | Rade2021Helper_R18_extra | 0.6022 | 0.6813 | 0.6276 |
| | | Rade2021Helper_ddpm | 0.6497 | 0.6887 | 0.6357 |
| | | Rade2021Helper_extra | 0.7034 | 0.8171 | 0.7049 |
| | | Rebuffi2021Fixing_106_16_cutmix_ddpm | 0.6285 | 0.6931 | 0.6197 |
| | | Rebuffi2021Fixing_28_10_cutmix_ddpm | 0.6632 | 0.7643 | 0.6584 |
| | | Rebuffi2021Fixing_70_16_cutmix_ddpm | 0.7032 | 0.7876 | 0.6797 |
| | | Rebuffi2021Fixing_70_16_cutmix_extra | 0.7857 | 0.8902 | 0.8009 |
| | | Rebuffi2021Fixing_70_16_cutmix_extra_L2 | 0.7857 | 0.8902 | 0.8009 |
| | | Rebuffi2021Fixing_70_16_cutmix_extra_Linf | 0.6582 | 0.8032 | 0.6929 |
| | | Rebuffi2021Fixing_R18_cutmix_ddpm | 0.6311 | 0.7067 | 0.6548 |
| | | Rebuffi2021Fixing_R18_ddpm | 0.5518 | 0.6140 | 0.5590 |
| | | Rice2020Overfitting | 0.6243 | 0.7074 | 0.5956 |
| | | Rony2019Decoupling | 0.6779 | 0.7773 | 0.6535 |
| | | Sehwag2020Hydra | 0.5841 | 0.6756 | 0.6009 |
| | | Sehwag2021Proxy | 0.6688 | 0.7279 | 0.6791 |
| | | Sehwag2021Proxy_R18 | 0.6730 | 0.7021 | 0.6586 |
| | | Sehwag2021Proxy_ResNest152 | 0.6098 | 0.6445 | 0.5890 |
| | | Sitawarin2020Improving | 0.5501 | 0.6599 | 0.6011 |
| | | Sridhar2021Robust | 0.5961 | 0.6798 | 0.6345 |
| | | Sridhar2021Robust_34_15 | 0.6059 | 0.6931 | 0.6179 |
| | | Standard | 0.3487 | 0.8632 | 0.6693 |
| | | Wang2020Improving | 0.5952 | 0.6722 | 0.6002 |
| | | Wang2023Better_WRN-28-10 | 0.7832 | 0.8562 | 0.7968 |
| | | Wang2023Better_WRN-70-16 | 0.7837 | 0.8560 | 0.7881 |
| | | Wong2020Fast | 0.5931 | 0.6554 | 0.5630 |
| | | Wu2020Adversarial | 0.6630 | 0.7294 | 0.6588 |
| | | Wu2020Adversarial_extra | 0.5805 | 0.6305 | 0.5964 |
| | | Xu2023Exploring_WRN-28-10 | 0.7296 | 0.8124 | 0.7378 |
| | | Zhang2019Theoretically | 0.5509 | 0.6169 | 0.5499 |
| | | Zhang2019You | 0.6186 | 0.6466 | 0.6169 |
| | | Zhang2020Attacks | 0.6206 | 0.6867 | 0.5899 |
| | | Zhang2020Geometry | 0.6227 | 0.7257 | 0.6638 |

Table 6. All CIFAR-10 accuracy scores for the Blur distortion and the victim model Inception-V3.

| Distortion | Victim Model | Dataset Model ID | **Ours** | CIFAR-10 | CIFAR-10-C |
|---|---|---|---|---|---|
| Blur | ResNet-50 | Addepalli2021Towards_RN18 | 0.5787 | 0.6462 | 0.6403 |
| | | AGAT | 0.2843 | 0.6679 | 0.6725 |
| | | Addepalli2021Towards_WRN34 | 0.6462 | 0.7154 | 0.6770 |
| | | Addepalli2022Efficient_RN18 | 0.6294 | 0.6822 | 0.6604 |
| | | Addepalli2022Efficient_WRN_34_10 | 0.6741 | 0.7707 | 0.6992 |
| | | Andriushchenko2020Understanding | 0.5427 | 0.6163 | 0.5930 |
| | | Augustin2020Adversarial | 0.6715 | 0.7490 | 0.7106 |
| | | Augustin2020Adversarial_34_10 | 0.6981 | 0.7928 | 0.7301 |
| | | Augustin2020Adversarial_34_10_extra | 0.7174 | 0.8186 | 0.7798 |
| | | Carmon2019Unlabeled | 0.6888 | 0.7757 | 0.7336 |
| | | Chen2020Adversarial | 0.6452 | 0.6590 | 0.6981 |
| | | Chen2020Efficient | 0.6275 | 0.6525 | 0.6725 |
| | | Chen2021LTD_WRN34_10 | 0.6335 | 0.6590 | 0.6433 |
| | | Chen2021LTD_WRN34_20 | 0.6770 | 0.7485 | 0.6880 |
| | | Cui2020Learnable_34_10 | 0.6551 | 0.7490 | 0.6804 |
| | | Cui2020Learnable_34_20 | 0.5978 | 0.7309 | 0.6874 |
| | | Dai2021Parameterizing | 0.6247 | 0.7609 | 0.7237 |
| | | Debenedetti2022Light_XCiT-L12 | 0.6981 | 0.8271 | 0.7362 |
| | | Debenedetti2022Light_XCiT-M12 | 0.7070 | 0.8008 | 0.7332 |
| | | Debenedetti2022Light_XCiT-S12 | 0.6837 | 0.7475 | 0.7023 |
| | | Diffenderfer2021Winning_Binary | 0.4640 | 0.8331 | 0.7926 |
| | | Diffenderfer2021Winning_Binary_CARD_Deck | 0.5066 | 0.8334 | 0.8310 |
| | | Diffenderfer2021Winning_LRR | 0.5968 | 0.9364 | 0.9066 |
| | | Diffenderfer2021Winning_LRR_CARD_Deck | 0.6214 | 0.8914 | 0.8589 |
| | | Ding2020MMA | 0.7105 | 0.7475 | 0.7248 |
| | | Engstrom2019Robustness | 0.7059 | 0.8071 | 0.7588 |
| | | Gowal2020Uncovering | 0.6921 | 0.7851 | 0.7473 |
| | | Gowal2020Uncovering_28_10_extra | 0.7043 | 0.7828 | 0.7314 |
| | | Gowal2020Uncovering_34_20 | 0.6239 | 0.6977 | 0.6705 |
| | | Gowal2020Uncovering_70_16 | 0.5895 | 0.6824 | 0.6659 |
| | | Gowal2020Uncovering_70_16_extra | 0.6673 | 0.7949 | 0.7340 |
| | | Gowal2020Uncovering_extra | 0.7446 | 0.8397 | 0.8021 |
| | | Gowal2021Improving_28_10_ddpm_100m | 0.6372 | 0.7531 | 0.7049 |
| | | Gowal2021Improving_70_16_ddpm_100m | 0.6492 | 0.7795 | 0.7472 |
| | | Gowal2021Improving_R18_ddpm_100m | 0.6448 | 0.7413 | 0.7013 |
| | | Hendrycks2019Using | 0.6047 | 0.6980 | 0.6467 |
| | | Hendrycks2020AugMix_ResNeXt | 0.5290 | 0.8900 | 0.8686 |
| | | Hendrycks2020AugMix_WRN | 0.4582 | 0.8427 | 0.8323 |
| | | Huang2020Self | 0.5931 | 0.7307 | 0.6873 |
| | | Huang2021Exploring | 0.7064 | 0.7769 | 0.7414 |
| | | Huang2021Exploring_ema | 0.7144 | 0.8171 | 0.7735 |
| | | Huang2022Revisiting_WRN-A4 | 0.6876 | 0.8077 | 0.7603 |
| | | Jia2022LAS-AT_34_10 | 0.5886 | 0.6375 | 0.6137 |
| | | Jia2022LAS-AT_70_16 | 0.6714 | 0.7642 | 0.7025 |
| | | Kang2021Stable | 0.7477 | 0.8424 | 0.7965 |
| | | Kireev2021Effectiveness_AugMixNoJSD | 0.4208 | 0.9261 | 0.8850 |
| | | Kireev2021Effectiveness_Gauss50percent | 0.4967 | 0.8489 | 0.7755 |
| | | Kireev2021Effectiveness_RLAT | 0.5867 | 0.7970 | 0.7477 |
| | | Kireev2021Effectiveness_RLATAugMix | 0.5535 | 0.8804 | 0.8560 |
| | | Kireev2021Effectiveness_RLATAugMixNoJSD | 0.4995 | 0.9102 | 0.8807 |
| | | Modas2021PRIMEResNet18 | 0.5925 | 0.7984 | 0.7706 |
| | | Pang2020Boosting | 0.5997 | 0.6891 | 0.6460 |
| | | Pang2022Robustness_WRN28_10 | 0.6289 | 0.7560 | 0.6996 |
| | | Pang2022Robustness_WRN70_16 | 0.6569 | 0.7384 | 0.6960 |
| | | Rade2021Helper_R18_ddpm | 0.6901 | 0.7827 | 0.7440 |
| | | Rade2021Helper_R18_extra | 0.6695 | 0.7727 | 0.7273 |
| | | Rade2021Helper_ddpm | 0.6564 | 0.7796 | 0.7247 |
| | | Rade2021Helper_extra | 0.7126 | 0.8112 | 0.7597 |
| | | Rebuffi2021Fixing_106_16_cutmix_ddpm | 0.6159 | 0.7851 | 0.7152 |
| | | Rebuffi2021Fixing_28_10_cutmix_ddpm | 0.6900 | 0.8035 | 0.7405 |
| | | Rebuffi2021Fixing_70_16_cutmix_ddpm | 0.7112 | 0.8332 | 0.7755 |
| | | Rebuffi2021Fixing_70_16_cutmix_extra | 0.7386 | 0.8628 | 0.8205 |
| | | Rebuffi2021Fixing_70_16_cutmix_extra_L2 | 0.7386 | 0.8628 | 0.8205 |
| | | Rebuffi2021Fixing_70_16_cutmix_extra_Linf | 0.6931 | 0.8055 | 0.7644 |
| | | Rebuffi2021Fixing_R18_cutmix_ddpm | 0.6588 | 0.7526 | 0.7133 |
| | | Rebuffi2021Fixing_R18_ddpm | 0.5824 | 0.7292 | 0.6685 |
| | | Rice2020Overfitting | 0.6355 | 0.7161 | 0.6890 |
| | | Rony2019Decoupling | 0.6919 | 0.7980 | 0.7377 |
| | | Sehwag2020Hydra | 0.6872 | 0.7389 | 0.7033 |
| | | Sehwag2021Proxy | 0.6722 | 0.7739 | 0.7254 |
| | | Sehwag2021Proxy_R18 | 0.6674 | 0.7341 | 0.6976 |
| | | Sehwag2021Proxy_ResNest152 | 0.6405 | 0.7231 | 0.6984 |
| | | Sitawarin2020Improving | 0.6529 | 0.7742 | 0.7014 |
| | | Sridhar2021Robust | 0.6850 | 0.7685 | 0.7329 |
| | | Sridhar2021Robust_34_15 | 0.6398 | 0.7593 | 0.7041 |
| | | Standard | 0.2418 | 0.8078 | 0.7419 |
| | | Wang2020Improving | 0.6361 | 0.7356 | 0.6852 |
| | | Wang2023Better_WRN-28-10 | 0.7341 | 0.8520 | 0.8157 |
| | | Wang2023Better_WRN-70-16 | 0.7449 | 0.8555 | 0.7941 |
| | | Wong2020Fast | 0.6208 | 0.6626 | 0.6458 |
| | | Wu2020Adversarial | 0.6597 | 0.7541 | 0.7046 |
| | | Wu2020Adversarial_extra | 0.6621 | 0.7926 | 0.7263 |
| | | Xu2023Exploring_WRN-28-10 | 0.7053 | 0.7963 | 0.7625 |
| | | Zhang2019Theoretically | 0.6080 | 0.6830 | 0.6418 |
| | | Zhang2019You | 0.6373 | 0.6896 | 0.6554 |
| | | Zhang2020Attacks | 0.6029 | 0.7498 | 0.7136 |
| | | Zhang2020Geometry | 0.7156 | 0.7993 | 0.7418 |

Table 7. All CIFAR-10 accuracy scores for the Blur distortion and the victim model ResNet-50.

| Distortion | Victim Model | Dataset Model ID | **Ours** | CIFAR-10 | CIFAR-10-C |
|---|---|---|---|---|---|
| Blur | VGG-16 | Addepalli2021Towards_RN18 | 0.5412 | 0.6179 | 0.5658 |
| | | AGAT | 0.2374 | 0.6565 | 0.6120 |
| | | Addepalli2021Towards_WRN34 | 0.5889 | 0.6795 | 0.6195 |
| | | Addepalli2022Efficient_RN18 | 0.5944 | 0.6602 | 0.6014 |
| | | Addepalli2022Efficient_WRN_34_10 | 0.6192 | 0.7093 | 0.6444 |
| | | Andriushchenko2020Understanding | 0.5234 | 0.5857 | 0.5236 |
| | | Augustin2020Adversarial | 0.6651 | 0.7670 | 0.7064 |
| | | Augustin2020Adversarial_34_10 | 0.7191 | 0.7910 | 0.7405 |
| | | Augustin2020Adversarial_34_10_extra | 0.7601 | 0.8733 | 0.8176 |
| | | Carmon2019Unlabeled | 0.6636 | 0.7521 | 0.6832 |
| | | Chen2020Adversarial | 0.5877 | 0.6439 | 0.5914 |
| | | Chen2020Efficient | 0.5809 | 0.6484 | 0.6043 |
| | | Chen2021LTD_WRN34_10 | 0.5955 | 0.6613 | 0.5980 |
| | | Chen2021LTD_WRN34_20 | 0.5780 | 0.6561 | 0.5975 |
| | | Cui2020Learnable_34_10 | 0.6285 | 0.7132 | 0.6462 |
| | | Cui2020Learnable_34_20 | 0.6121 | 0.6906 | 0.6244 |
| | | Dai2021Parameterizing | 0.6021 | 0.7002 | 0.6271 |
| | | Debenedetti2022Light_XCiT-L12 | 0.7209 | 0.7690 | 0.7265 |
| | | Debenedetti2022Light_XCiT-M12 | 0.7160 | 0.8015 | 0.7144 |
| | | Debenedetti2022Light_XCiT-S12 | 0.6590 | 0.7338 | 0.6808 |
| | | Diffenderfer2021Winning_Binary | 0.5221 | 0.9024 | 0.8712 |
| | | Diffenderfer2021Winning_Binary_CARD_Deck | 0.5582 | 0.9030 | 0.8358 |
| | | Diffenderfer2021Winning_LRR | 0.5413 | 0.9738 | 0.9305 |
| | | Diffenderfer2021Winning_LRR_CARD_Deck | 0.5924 | 0.9477 | 0.9019 |
| | | Ding2020MMA | 0.6053 | 0.6799 | 0.6201 |
| | | Engstrom2019Robustness | 0.7022 | 0.7749 | 0.6918 |
| | | Gowal2020Uncovering | 0.6544 | 0.7426 | 0.6793 |
| | | Gowal2020Uncovering_28_10_extra | 0.6561 | 0.7306 | 0.6747 |
| | | Gowal2020Uncovering_34_20 | 0.5833 | 0.6442 | 0.5966 |
| | | Gowal2020Uncovering_70_16 | 0.5659 | 0.6493 | 0.5830 |
| | | Gowal2020Uncovering_70_16_extra | 0.6999 | 0.7822 | 0.7219 |
| | | Gowal2020Uncovering_extra | 0.7721 | 0.8843 | 0.8172 |
| | | Gowal2021Improving_28_10_ddpm_100m | 0.6478 | 0.7092 | 0.6608 |
| | | Gowal2021Improving_70_16_ddpm_100m | 0.6522 | 0.7375 | 0.6806 |
| | | Gowal2021Improving_R18_ddpm_100m | 0.6089 | 0.6802 | 0.6371 |
| | | Hendrycks2019Using | 0.6072 | 0.6930 | 0.6272 |
| | | Hendrycks2020AugMix_ResNeXt | 0.5406 | 0.8975 | 0.8630 |
| | | Hendrycks2020AugMix_WRN | 0.4770 | 0.8780 | 0.8525 |
| | | Huang2020Self | 0.5572 | 0.6062 | 0.5613 |
| | | Huang2021Exploring | 0.6637 | 0.7647 | 0.7051 |
| | | Huang2021Exploring_ema | 0.7054 | 0.8025 | 0.7271 |
| | | Huang2022Revisiting_WRN-A4 | 0.7012 | 0.8210 | 0.7471 |
| | | Jia2022LAS-AT_34_10 | 0.5871 | 0.6703 | 0.6058 |
| | | Jia2022LAS-AT_70_16 | 0.5973 | 0.6705 | 0.6055 |
| | | Kang2021Stable | 0.7573 | 0.8689 | 0.7857 |
| | | Kireev2021Effectiveness_AugMixNoJSD | 0.3979 | 0.8829 | 0.8634 |
| | | Kireev2021Effectiveness_Gauss50percent | 0.4847 | 0.7835 | 0.6981 |
| | | Kireev2021Effectiveness_RLAT | 0.5897 | 0.7663 | 0.6966 |
| | | Kireev2021Effectiveness_RLATAugMix | 0.5559 | 0.8902 | 0.8433 |
| | | Kireev2021Effectiveness_RLATAugMixNoJSD | 0.5609 | 0.9076 | 0.8846 |
| | | Modas2021PRIMEResNet18 | 0.6117 | 0.8340 | 0.7868 |
| | | Pang2020Boosting | 0.5760 | 0.6680 | 0.6186 |
| | | Pang2022Robustness_WRN28_10 | 0.6143 | 0.6813 | 0.6170 |
| | | Pang2022Robustness_WRN70_16 | 0.6177 | 0.6862 | 0.6357 |
| | | Rade2021Helper_R18_ddpm | 0.6942 | 0.7941 | 0.7122 |
| | | Rade2021Helper_R18_extra | 0.6698 | 0.7373 | 0.6841 |
| | | Rade2021Helper_ddpm | 0.6293 | 0.6954 | 0.6279 |
| | | Rade2021Helper_extra | 0.7028 | 0.8141 | 0.7337 |
| | | Rebuffi2021Fixing_106_16_cutmix_ddpm | 0.6341 | 0.7082 | 0.6471 |
| | | Rebuffi2021Fixing_28_10_cutmix_ddpm | 0.6941 | 0.7918 | 0.7372 |
| | | Rebuffi2021Fixing_70_16_cutmix_ddpm | 0.7128 | 0.8288 | 0.7487 |
| | | Rebuffi2021Fixing_70_16_cutmix_extra | 0.7712 | 0.8946 | 0.8378 |
| | | Rebuffi2021Fixing_70_16_cutmix_extra_L2 | 0.7712 | 0.8946 | 0.8378 |
| | | Rebuffi2021Fixing_70_16_cutmix_extra_Linf | 0.6976 | 0.8036 | 0.7351 |
| | | Rebuffi2021Fixing_R18_cutmix_ddpm | 0.6510 | 0.7524 | 0.6823 |
| | | Rebuffi2021Fixing_R18_ddpm | 0.5713 | 0.6213 | 0.5761 |
| | | Rice2020Overfitting | 0.6150 | 0.6795 | 0.6213 |
| | | Rony2019Decoupling | 0.6693 | 0.7484 | 0.6331 |
| | | Sehwag2020Hydra | 0.6492 | 0.7318 | 0.6662 |
| | | Sehwag2021Proxy | 0.6777 | 0.7707 | 0.7200 |
| | | Sehwag2021Proxy_R18 | 0.6349 | 0.7342 | 0.6901 |
| | | Sehwag2021Proxy_ResNest152 | 0.6273 | 0.7121 | 0.6473 |
| | | Sitawarin2020Improving | 0.6099 | 0.6998 | 0.6209 |
| | | Sridhar2021Robust | 0.6558 | 0.7387 | 0.6777 |
| | | Sridhar2021Robust_34_15 | 0.6334 | 0.7243 | 0.6520 |
| | | Standard | 0.3404 | 0.9069 | 0.6913 |
| | | Wang2020Improving | 0.6427 | 0.7102 | 0.6592 |
| | | Wang2023Better_WRN-28-10 | 0.7675 | 0.8816 | 0.8151 |
| | | Wang2023Better_WRN-70-16 | 0.7811 | 0.8964 | 0.8355 |
| | | Wong2020Fast | 0.5707 | 0.6355 | 0.5485 |
| | | Wu2020Adversarial | 0.6445 | 0.7189 | 0.6348 |
| | | Wu2020Adversarial_extra | 0.6470 | 0.7191 | 0.6573 |
| | | Xu2023Exploring_WRN-28-10 | 0.6971 | 0.8504 | 0.7632 |
| | | Zhang2019Theoretically | 0.5936 | 0.6719 | 0.6101 |
| | | Zhang2019You | 0.6417 | 0.6987 | 0.6370 |
| | | Zhang2020Attacks | 0.5806 | 0.6692 | 0.6049 |
| | | Zhang2020Geometry | 0.6456 | 0.7567 | 0.6814 |

Table 8. All CIFAR-10 accuracy scores for the Blur distortion and the victim model VGG-16.

| Distortion | Victim Model | Dataset Model ID | **Ours** | CIFAR-10 | CIFAR-10-C |
|---|---|---|---|---|---|
| Noise | Inception-V3 | Addepalli2021Towards_RN18 | 0.5686 | 0.8136 | 0.7618 |
| | | AGAT | 0.3567 | 0.8616 | 0.6471 |
| | | Addepalli2021Towards_WRN34 | 0.5099 | 0.8475 | 0.7625 |
| | | Addepalli2022Efficient_RN18 | 0.5910 | 0.8551 | 0.8085 |
| | | Addepalli2022Efficient_WRN_34_10 | 0.6364 | 0.8839 | 0.8337 |
| | | Andriushchenko2020Understanding | 0.5323 | 0.7919 | 0.7497 |
| | | Augustin2020Adversarial | 0.5555 | 0.9222 | 0.7916 |
| | | Augustin2020Adversarial_34_10 | 0.5935 | 0.9295 | 0.8258 |
| | | Augustin2020Adversarial_34_10_extra | 0.6076 | 0.9468 | 0.8788 |
| | | Carmon2019Unlabeled | 0.6106 | 0.8994 | 0.8217 |
| | | Chen2020Adversarial | 0.5499 | 0.8513 | 0.7972 |
| | | Chen2020Efficient | 0.5917 | 0.8473 | 0.7938 |
| | | Chen2021LTD_WRN34_10 | 0.6070 | 0.8512 | 0.7947 |
| | | Chen2021LTD_WRN34_20 | 0.5998 | 0.8599 | 0.7988 |
| | | Cui2020Learnable_34_10 | 0.6292 | 0.8799 | 0.8233 |
| | | Cui2020Learnable_34_20 | 0.5917 | 0.8792 | 0.8267 |
| | | Dai2021Parameterizing | 0.6290 | 0.8675 | 0.8059 |
| | | Debenedetti2022Light_XCiT-L12 | 0.6413 | 0.9106 | 0.8467 |
| | | Debenedetti2022Light_XCiT-M12 | 0.6241 | 0.9062 | 0.8334 |
| | | Debenedetti2022Light_XCiT-S12 | 0.6065 | 0.8929 | 0.8356 |
| | | Diffenderfer2021Winning_Binary | 0.5265 | 0.9696 | 0.8359 |
| | | Diffenderfer2021Winning_Binary_CARD_Deck | 0.6777 | 0.9706 | 0.9269 |
| | | Diffenderfer2021Winning_LRR | 0.5475 | 0.9832 | 0.8098 |
| | | Diffenderfer2021Winning_LRR_CARD_Deck | 0.6867 | 0.9841 | 0.9337 |
| | | Ding2020MMA | 0.6246 | 0.8872 | 0.8382 |
| | | Engstrom2019Robustness | 0.6445 | 0.9213 | 0.8651 |
| | | Gowal2020Uncovering | 0.7190 | 0.9226 | 0.8875 |
| | | Gowal2020Uncovering_28_10_extra | 0.6071 | 0.9006 | 0.8292 |
| | | Gowal2020Uncovering_34_20 | 0.6049 | 0.8544 | 0.8085 |
| | | Gowal2020Uncovering_70_16 | 0.5978 | 0.8543 | 0.8087 |
| | | Gowal2020Uncovering_70_16_extra | 0.6282 | 0.9153 | 0.8358 |
| | | Gowal2020Uncovering_extra | 0.6189 | 0.9637 | 0.8878 |
| | | Gowal2021Improving_28_10_ddpm_100m | 0.5713 | 0.8816 | 0.8022 |
| | | Gowal2021Improving_70_16_ddpm_100m | 0.5980 | 0.8880 | 0.8050 |
| | | Gowal2021Improving_R18_ddpm_100m | 0.5971 | 0.8724 | 0.8246 |
| | | Hendrycks2019Using | 0.5366 | 0.8728 | 0.7805 |
| | | Hendrycks2020AugMix_ResNeXt | 0.5499 | 0.9724 | 0.7871 |
| | | Hendrycks2020AugMix_WRN | 0.5472 | 0.9710 | 0.8170 |
| | | Huang2020Self | 0.5872 | 0.8309 | 0.7837 |
| | | Huang2021Exploring | 0.6494 | 0.9108 | 0.8263 |
| | | Huang2021Exploring_ema | 0.6144 | 0.9225 | 0.8384 |
| | | Huang2022Revisiting_WRN-A4 | 0.6064 | 0.9207 | 0.8376 |
| | | Jia2022LAS-AT_34_10 | 0.6006 | 0.8473 | 0.7883 |
| | | Jia2022LAS-AT_70_16 | 0.5886 | 0.8544 | 0.7973 |
| | | Kang2021Stable | 0.5620 | 0.9458 | 0.8631 |
| | | Kireev2021Effectiveness_AugMixNoJSD | 0.5093 | 0.9701 | 0.7754 |
| | | Kireev2021Effectiveness_Gauss50percent | 0.6736 | 0.9483 | 0.9279 |
| | | Kireev2021Effectiveness_RLAT | 0.5774 | 0.9446 | 0.8718 |
| | | Kireev2021Effectiveness_RLATAugMix | 0.6016 | 0.9643 | 0.8715 |
| | | Kireev2021Effectiveness_RLATAugMixNoJSD | 0.5980 | 0.9727 | 0.8620 |
| | | Modas2021PRIMEResNet18 | 0.8057 | 0.9357 | 0.8833 |
| | | Pang2020Boosting | 0.5679 | 0.8485 | 0.7764 |
| | | Pang2022Robustness_WRN28_10 | 0.5754 | 0.8875 | 0.8291 |
| | | Pang2022Robustness_WRN70_16 | 0.5871 | 0.8855 | 0.8265 |
| | | Rade2021Helper_R18_ddpm | 0.6644 | 0.9185 | 0.8693 |
| | | Rade2021Helper_R18_extra | 0.4346 | 0.8906 | 0.8135 |
| | | Rade2021Helper_ddpm | 0.5785 | 0.8838 | 0.7837 |
| | | Rade2021Helper_extra | 0.6232 | 0.9154 | 0.8456 |
| | | Rebuffi2021Fixing_106_16_cutmix_ddpm | 0.5645 | 0.8832 | 0.8121 |
| | | Rebuffi2021Fixing_28_10_cutmix_ddpm | 0.5947 | 0.9197 | 0.8632 |
| | | Rebuffi2021Fixing_70_16_cutmix_ddpm | 0.5582 | 0.8872 | 0.8157 |
| | | Rebuffi2021Fixing_70_16_cutmix_extra | 0.5702 | 0.9286 | 0.8325 |
| | | Rebuffi2021Fixing_70_16_cutmix_extra_L2 | 0.5665 | 0.9654 | 0.8615 |
| | | Rebuffi2021Fixing_70_16_cutmix_extra_Linf | 0.5702 | 0.9286 | 0.8325 |
| | | Rebuffi2021Fixing_R18_cutmix_ddpm | 0.5947 | 0.9062 | 0.8400 |
| | | Rebuffi2021Fixing_R18_ddpm | 0.5782 | 0.8316 | 0.7772 |
| | | Rice2020Overfitting | 0.6384 | 0.8896 | 0.8373 |
| | | Rony2019Decoupling | 0.6673 | 0.9043 | 0.8789 |
| | | Sehwag2020Hydra | 0.5996 | 0.8845 | 0.8265 |
| | | Sehwag2021Proxy | 0.6686 | 0.9082 | 0.8530 |
| | | Sehwag2021Proxy_R18 | 0.6419 | 0.8992 | 0.8469 |
| | | Sehwag2021Proxy_ResNest152 | 0.6088 | 0.8766 | 0.8039 |
| | | Sitawarin2020Improving | 0.5854 | 0.8750 | 0.8235 |
| | | Sridhar2021Robust | 0.6082 | 0.8964 | 0.8104 |
| | | Sridhar2021Robust_34_15 | 0.5891 | 0.8665 | 0.7879 |
| | | Standard | 0.3344 | 0.9745 | 0.4484 |
| | | Wang2020Improving | 0.5869 | 0.8763 | 0.7961 |
| | | Wang2023Better_WRN-28-10 | 0.6408 | 0.9292 | 0.8574 |
| | | Wang2023Better_WRN-70-16 | 0.6428 | 0.9661 | 0.8814 |
| | | Wong2020Fast | 0.5003 | 0.8348 | 0.7665 |
| | | Wu2020Adversarial | 0.6676 | 0.8828 | 0.8562 |
| | | Wu2020Adversarial_extra | 0.6109 | 0.8838 | 0.7977 |
| | | Xu2023Exploring_WRN-28-10 | 0.6000 | 0.9417 | 0.8705 |
| | | Zhang2019Theoretically | 0.5925 | 0.8525 | 0.7796 |
| | | Zhang2019You | 0.5869 | 0.8645 | 0.8390 |
| | | Zhang2020Attacks | 0.5745 | 0.8471 | 0.7992 |
| | | Zhang2020Geometry | 0.5815 | 0.8933 | 0.8143 |

Table 9. All CIFAR-10 accuracy scores for the Noise distortion and the victim model Inception-V3.

| Distortion | Victim Model | Dataset Model ID | **Ours** | CIFAR-10 | CIFAR-10-C |
|---|---|---|---|---|---|
| Noise | ResNet-50 | Addepalli2021Towards_RN18 | 0.5987 | 0.8174 | 0.7752 |
| | | AGAT | 0.3403 | 0.8470 | 0.6424 |
| | | Addepalli2021Towards_WRN34 | 0.6165 | 0.8704 | 0.7874 |
| | | Addepalli2022Efficient_RN18 | 0.6528 | 0.8820 | 0.8308 |
| | | Addepalli2022Efficient_WRN_34_10 | 0.7058 | 0.9052 | 0.8541 |
| | | Andriushchenko2020Understanding | 0.5710 | 0.8099 | 0.7705 |
| | | Augustin2020Adversarial | 0.6325 | 0.9323 | 0.8133 |
| | | Augustin2020Adversarial_34_10 | 0.6765 | 0.9418 | 0.8411 |
| | | Augustin2020Adversarial_34_10_extra | 0.6703 | 0.9569 | 0.8896 |
| | | Carmon2019Unlabeled | 0.6728 | 0.9122 | 0.8409 |
| | | Chen2020Adversarial | 0.6039 | 0.8766 | 0.8244 |
| | | Chen2020Efficient | 0.6371 | 0.8717 | 0.8162 |
| | | Chen2021LTD_WRN34_10 | 0.6340 | 0.8702 | 0.8135 |
| | | Chen2021LTD_WRN34_20 | 0.6323 | 0.8789 | 0.8239 |
| | | Cui2020Learnable_34_10 | 0.6620 | 0.9016 | 0.8449 |
| | | Cui2020Learnable_34_20 | 0.6486 | 0.9022 | 0.8463 |
| | | Dai2021Parameterizing | 0.6713 | 0.8887 | 0.8295 |
| | | Debenedetti2022Light_XCiT-L12 | 0.7140 | 0.9301 | 0.8652 |
| | | Debenedetti2022Light_XCiT-M12 | 0.6939 | 0.9276 | 0.8570 |
| | | Debenedetti2022Light_XCiT-S12 | 0.6629 | 0.9161 | 0.8542 |
| | | Diffenderfer2021Winning_Binary | 0.5368 | 0.9740 | 0.8550 |
| | | Diffenderfer2021Winning_Binary_CARD_Deck | 0.6605 | 0.9782 | 0.9410 |
| | | Diffenderfer2021Winning_LRR | 0.6402 | 0.9877 | 0.8214 |
| | | Diffenderfer2021Winning_LRR_CARD_Deck | 0.7214 | 0.9858 | 0.9437 |
| | | Ding2020MMA | 0.6972 | 0.8988 | 0.8511 |
| | | Engstrom2019Robustness | 0.6839 | 0.9307 | 0.8813 |
| | | Gowal2020Uncovering | 0.7776 | 0.9304 | 0.8982 |
| | | Gowal2020Uncovering_28_10_extra | 0.6742 | 0.9140 | 0.8469 |
| | | Gowal2020Uncovering_34_20 | 0.6571 | 0.8755 | 0.8325 |
| | | Gowal2020Uncovering_70_16 | 0.6371 | 0.8719 | 0.8313 |
| | | Gowal2020Uncovering_70_16_extra | 0.6971 | 0.9262 | 0.8537 |
| | | Gowal2020Uncovering_extra | 0.6859 | 0.9642 | 0.9003 |
| | | Gowal2021Improving_28_10_ddpm_100m | 0.6490 | 0.8960 | 0.8262 |
| | | Gowal2021Improving_70_16_ddpm_100m | 0.6593 | 0.9095 | 0.8311 |
| | | Gowal2021Improving_R18_ddpm_100m | 0.6525 | 0.8883 | 0.8401 |
| | | Hendrycks2019Using | 0.5955 | 0.8888 | 0.8019 |
| | | Hendrycks2020AugMix_ResNeXt | 0.6072 | 0.9784 | 0.7904 |
| | | Hendrycks2020AugMix_WRN | 0.5855 | 0.9740 | 0.8238 |
| | | Huang2020Self | 0.6193 | 0.8505 | 0.8067 |
| | | Huang2021Exploring | 0.6878 | 0.9241 | 0.8432 |
| | | Huang2021Exploring_ema | 0.6856 | 0.9338 | 0.8536 |
| | | Huang2022Revisiting_WRN-A4 | 0.6797 | 0.9317 | 0.8549 |
| | | Jia2022LAS-AT_34_10 | 0.6395 | 0.8667 | 0.8086 |
| | | Jia2022LAS-AT_70_16 | 0.6417 | 0.8734 | 0.8231 |
| | | Kang2021Stable | 0.6223 | 0.9514 | 0.8780 |
| | | Kireev2021Effectiveness_AugMixNoJSD | 0.5075 | 0.9726 | 0.7818 |
| | | Kireev2021Effectiveness_Gauss50percent | 0.6403 | 0.9548 | 0.9341 |
| | | Kireev2021Effectiveness_RLAT | 0.6056 | 0.9524 | 0.8833 |
| | | Kireev2021Effectiveness_RLATAugMix | 0.6296 | 0.9715 | 0.8863 |
| | | Kireev2021Effectiveness_RLATAugMixNoJSD | 0.5820 | 0.9761 | 0.8751 |
| | | Modas2021PRIMEResNet18 | 0.8373 | 0.9549 | 0.9090 |
| | | Pang2020Boosting | 0.6165 | 0.8661 | 0.7945 |
| | | Pang2022Robustness_WRN28_10 | 0.6369 | 0.9061 | 0.8516 |
| | | Pang2022Robustness_WRN70_16 | 0.6445 | 0.9053 | 0.8497 |
| | | Rade2021Helper_R18_ddpm | 0.7358 | 0.9307 | 0.8811 |
| | | Rade2021Helper_R18_extra | 0.5438 | 0.9037 | 0.8322 |
| | | Rade2021Helper_ddpm | 0.6612 | 0.9002 | 0.7999 |
| | | Rade2021Helper_extra | 0.6816 | 0.9278 | 0.8627 |
| | | Rebuffi2021Fixing_106_16_cutmix_ddpm | 0.6313 | 0.9042 | 0.8364 |
| | | Rebuffi2021Fixing_28_10_cutmix_ddpm | 0.6628 | 0.9386 | 0.8815 |
| | | Rebuffi2021Fixing_70_16_cutmix_ddpm | 0.6158 | 0.9033 | 0.8389 |
| | | Rebuffi2021Fixing_70_16_cutmix_extra | 0.6471 | 0.9690 | 0.8739 |
| | | Rebuffi2021Fixing_70_16_cutmix_extra_L2 | 0.6471 | 0.9690 | 0.8739 |
| | | Rebuffi2021Fixing_70_16_cutmix_extra_Linf | 0.6431 | 0.9382 | 0.8503 |
| | | Rebuffi2021Fixing_R18_cutmix_ddpm | 0.6811 | 0.9212 | 0.8576 |
| | | Rebuffi2021Fixing_R18_ddpm | 0.6197 | 0.8557 | 0.7984 |
| | | Rice2020Overfitting | 0.6952 | 0.9043 | 0.8518 |
| | | Rony2019Decoupling | 0.7091 | 0.9098 | 0.8847 |
| | | Sehwag2020Hydra | 0.6723 | 0.9077 | 0.8466 |
| | | Sehwag2021Proxy | 0.7222 | 0.9276 | 0.8709 |
| | | Sehwag2021Proxy_R18 | 0.7019 | 0.9169 | 0.8616 |
| | | Sehwag2021Proxy_ResNest152 | 0.6812 | 0.8954 | 0.8276 |
| | | Sitawarin2020Improving | 0.6513 | 0.8874 | 0.8393 |
| | | Sridhar2021Robust | 0.6752 | 0.9131 | 0.8327 |
| | | Sridhar2021Robust_34_15 | 0.6420 | 0.8861 | 0.8066 |
| | | Standard | 0.3645 | 0.9708 | 0.4667 |
| | | Wang2020Improving | 0.6501 | 0.8891 | 0.8153 |
| | | Wang2023Better_WRN-28-10 | 0.6910 | 0.9398 | 0.8764 |
| | | Wang2023Better_WRN-70-16 | 0.7118 | 0.9478 | 0.8774 |
| | | Wong2020Fast | 0.5620 | 0.8467 | 0.7806 |
| | | Wu2020Adversarial | 0.7150 | 0.9030 | 0.8691 |
| | | Wu2020Adversarial_extra | 0.6646 | 0.9026 | 0.8182 |
| | | Xu2023Exploring_WRN-28-10 | 0.6841 | 0.9529 | 0.8869 |
| | | Zhang2019Theoretically | 0.6505 | 0.8703 | 0.8070 |
| | | Zhang2019You | 0.6472 | 0.8870 | 0.8586 |
| | | Zhang2020Attacks | 0.6145 | 0.8682 | 0.8166 |
| | | Zhang2020Geometry | 0.6526 | 0.9109 | 0.8316 |

Table 10. All CIFAR-10 accuracy scores for the Noise distortion and the victim model ResNet-50.

| Distortion | Victim Model | Dataset Model ID | **Ours** | CIFAR-10 | CIFAR-10-C |
|---|---|---|---|---|---|
| Noise | VGG-16 | Addepalli2021Towards_RN18 | 0.5032 | 0.7885 | 0.7383 |
| | | AGAT | 0.2747 | 0.8396 | 0.6173 |
| | | Addepalli2021Towards_WRN34 | 0.4837 | 0.8314 | 0.7422 |
| | | Addepalli2022Efficient_RN18 | 0.5281 | 0.8398 | 0.7899 |
| | | Addepalli2022Efficient_WRN_34_10 | 0.5810 | 0.8739 | 0.8173 |
| | | Andriushchenko2020Understanding | 0.4590 | 0.7731 | 0.7276 |
| | | Augustin2020Adversarial | 0.4960 | 0.9076 | 0.7703 |
| | | Augustin2020Adversarial_34_10 | 0.5438 | 0.9230 | 0.8092 |
| | | Augustin2020Adversarial_34_10_extra | 0.5480 | 0.9440 | 0.8688 |
| | | Carmon2019Unlabeled | 0.5347 | 0.8917 | 0.8059 |
| | | Chen2020Adversarial | 0.4795 | 0.8281 | 0.7701 |
| | | Chen2020Efficient | 0.5126 | 0.8307 | 0.7740 |
| | | Chen2021LTD_WRN34_10 | 0.5246 | 0.8310 | 0.7734 |
| | | Chen2021LTD_WRN34_20 | 0.5074 | 0.8432 | 0.7796 |
| | | Cui2020Learnable_34_10 | 0.5380 | 0.8740 | 0.8084 |
| | | Cui2020Learnable_34_20 | 0.5197 | 0.8656 | 0.8083 |
| | | Dai2021Parameterizing | 0.5525 | 0.8528 | 0.7875 |
| | | Debenedetti2022Light_XCiT-L12 | 0.5983 | 0.9046 | 0.8350 |
| | | Debenedetti2022Light_XCiT-M12 | 0.5891 | 0.9026 | 0.8249 |
| | | Debenedetti2022Light_XCiT-S12 | 0.5501 | 0.8851 | 0.8227 |
| | | Diffenderfer2021Winning_Binary | 0.4162 | 0.9693 | 0.8220 |
| | | Diffenderfer2021Winning_Binary_CARD_Deck | 0.5439 | 0.9727 | 0.9179 |
| | | Diffenderfer2021Winning_LRR | 0.5013 | 0.9885 | 0.7997 |
| | | Diffenderfer2021Winning_LRR_CARD_Deck | 0.6036 | 0.9868 | 0.9287 |
| | | Ding2020MMA | 0.5709 | 0.8708 | 0.8160 |
| | | Engstrom2019Robustness | 0.5803 | 0.9069 | 0.8481 |
| | | Gowal2020Uncovering | 0.6615 | 0.9083 | 0.8722 |
| | | Gowal2020Uncovering_28_10_extra | 0.5421 | 0.8935 | 0.8177 |
| | | Gowal2020Uncovering_34_20 | 0.5406 | 0.8415 | 0.7916 |
| | | Gowal2020Uncovering_70_16 | 0.5191 | 0.8392 | 0.7906 |
| | | Gowal2020Uncovering_70_16_extra | 0.5618 | 0.9085 | 0.8253 |
| | | Gowal2020Uncovering_extra | 0.5631 | 0.9566 | 0.8801 |
| | | Gowal2021Improving_28_10_ddpm_100m | 0.5102 | 0.8656 | 0.7819 |
| | | Gowal2021Improving_70_16_ddpm_100m | 0.5370 | 0.8720 | 0.7854 |
| | | Gowal2021Improving_R18_ddpm_100m | 0.5306 | 0.8537 | 0.7999 |
| | | Hendrycks2019Using | 0.4791 | 0.8598 | 0.7641 |
| | | Hendrycks2020AugMix_ResNeXt | 0.4902 | 0.9714 | 0.7719 |
| | | Hendrycks2020AugMix_WRN | 0.4321 | 0.9647 | 0.7947 |
| | | Huang2020Self | 0.5097 | 0.8137 | 0.7677 |
| | | Huang2021Exploring | 0.5617 | 0.9030 | 0.8124 |
| | | Huang2021Exploring_ema | 0.5506 | 0.9125 | 0.8258 |
| | | Huang2022Revisiting_WRN-A4 | 0.5476 | 0.9163 | 0.8247 |
| | | Jia2022LAS-AT_34_10 | 0.5171 | 0.8264 | 0.7678 |
| | | Jia2022LAS-AT_70_16 | 0.5208 | 0.8388 | 0.7792 |
| | | Kang2021Stable | 0.5015 | 0.9394 | 0.8511 |
| | | Kireev2021Effectiveness_AugMixNoJSD | 0.3938 | 0.9711 | 0.7575 |
| | | Kireev2021Effectiveness_Gauss50percent | 0.5638 | 0.9441 | 0.9199 |
| | | Kireev2021Effectiveness_RLAT | 0.5114 | 0.9409 | 0.8565 |
| | | Kireev2021Effectiveness_RLATAugMix | 0.4918 | 0.9635 | 0.8592 |
| | | Kireev2021Effectiveness_RLATAugMixNoJSD | 0.4707 | 0.9735 | 0.8508 |
| | | Modas2021PRIMEResNet18 | 0.7093 | 0.9283 | 0.8717 |
| | | Pang2020Boosting | 0.5028 | 0.8363 | 0.7551 |
| | | Pang2022Robustness_WRN28_10 | 0.5092 | 0.8691 | 0.8095 |
| | | Pang2022Robustness_WRN70_16 | 0.5106 | 0.8681 | 0.8062 |
| | | Rade2021Helper_R18_ddpm | 0.6194 | 0.9110 | 0.8575 |
| | | Rade2021Helper_R18_extra | 0.4385 | 0.8784 | 0.7991 |
| | | Rade2021Helper_ddpm | 0.5347 | 0.8685 | 0.7604 |
| | | Rade2021Helper_extra | 0.5613 | 0.9094 | 0.8311 |
| | | Rebuffi2021Fixing_106_16_cutmix_ddpm | 0.5111 | 0.8719 | 0.7952 |
| | | Rebuffi2021Fixing_28_10_cutmix_ddpm | 0.5463 | 0.9079 | 0.8515 |
| | | Rebuffi2021Fixing_70_16_cutmix_ddpm | 0.4920 | 0.8749 | 0.8001 |
| | | Rebuffi2021Fixing_70_16_cutmix_extra | 0.5225 | 0.9605 | 0.8520 |
| | | Rebuffi2021Fixing_70_16_cutmix_extra_L2 | 0.5225 | 0.9605 | 0.8520 |
| | | Rebuffi2021Fixing_70_16_cutmix_extra_Linf | 0.5047 | 0.9191 | 0.8190 |
| | | Rebuffi2021Fixing_R18_cutmix_ddpm | 0.5580 | 0.8915 | 0.8251 |
| | | Rebuffi2021Fixing_R18_ddpm | 0.5093 | 0.8133 | 0.7573 |
| | | Rice2020Overfitting | 0.5758 | 0.8678 | 0.8173 |
| | | Rony2019Decoupling | 0.5889 | 0.8985 | 0.8656 |
| | | Sehwag2020Hydra | 0.5441 | 0.8718 | 0.8104 |
| | | Sehwag2021Proxy | 0.5923 | 0.8937 | 0.8360 |
| | | Sehwag2021Proxy_R18 | 0.5687 | 0.8839 | 0.8316 |
| | | Sehwag2021Proxy_ResNest152 | 0.5531 | 0.8587 | 0.7847 |
| | | Sitawarin2020Improving | 0.5113 | 0.8612 | 0.8043 |
| | | Sridhar2021Robust | 0.5443 | 0.8887 | 0.7959 |
| | | Sridhar2021Robust_34_15 | 0.5247 | 0.8558 | 0.7708 |
| | | Standard | 0.2984 | 0.9713 | 0.4432 |
| | | Wang2020Improving | 0.5365 | 0.8621 | 0.7723 |
| | | Wang2023Better_WRN-28-10 | 0.5613 | 0.9187 | 0.8421 |
| | | Wang2023Better_WRN-70-16 | 0.5786 | 0.9603 | 0.8738 |
| | | Wong2020Fast | 0.4447 | 0.8189 | 0.7458 |
| | | Wu2020Adversarial | 0.6023 | 0.8640 | 0.8377 |
| | | Wu2020Adversarial_extra | 0.5333 | 0.8746 | 0.7786 |
| | | Xu2023Exploring_WRN-28-10 | 0.5556 | 0.9329 | 0.8545 |
| | | Zhang2019Theoretically | 0.5145 | 0.8328 | 0.7588 |
| | | Zhang2019You | 0.5182 | 0.8453 | 0.8175 |
| | | Zhang2020Attacks | 0.5097 | 0.8296 | 0.7793 |
| | | Zhang2020Geometry | 0.5153 | 0.8828 | 0.7961 |

Table 11. All CIFAR-10 accuracy scores for the Noise distortion and the victim model VGG-16.

## References

[1] Sara Beery, Elijah Cole, and Arvi Gjoka. The iwildcam 2020 competition dataset. *arXiv preprint arXiv:2004.10340*, 2020. 2

[2] Nicholas Carlini, Anish Athalye, Nicolas Papernot, Wieland Brendel, Jonas Rauber, Dimitris Tsipras, Ian Goodfellow, Aleksander Madry, and Alexey Kurakin. On evaluating adversarial robustness. *arXiv preprint arXiv:1902.06705*, 2019. 5, 9

[3] Jeremy Cohen, Elan Rosenfeld, and Zico Kolter. Certified adversarial robustness via randomized smoothing. In *international conference on machine learning*, pages 1310–1320. PMLR, 2019. 5, 9

[4] Francesco Croce, Maksym Andriushchenko, Vikash Sehwag, Edoardo Debenedetti, Nicolas Flammarion, Mung Chiang, Prateek Mittal, and Matthias Hein. RobustBench: a standardized adversarial robustness benchmark. In *Thirty-fifth Conference on Neural Information Processing Systems Datasets and Benchmarks Track*, pages 1–17, 2021. 5, 6, 9, 12, 13

[5] Francesco Croce and Matthias Hein. Reliable evaluation of adversarial robustness with an ensemble of diverse parameter-free attacks. In *International conference on machine learning*, pages 2206–2216. PMLR, 2020. 2

[6] Ekin D Cubuk, Barret Zoph, Dandelion Mane, Vijay Vasudevan, and Quoc V Le. Autoaugment: Learning augmentation policies from data. *arXiv preprint arXiv:1805.09501*, 2018. 2

[7] Ekin D Cubuk, Barret Zoph, Jonathon Shlens, and Quoc V Le. Randaugment: Practical automated data augmentation with a reduced search space. In *Proceedings of the IEEE/CVF conference on computer vision and pattern recognition workshops*, pages 702–703, 2020. 2

[8] Jia Deng, Wei Dong, Richard Socher, Li-Jia Li, Kai Li, and Li Fei-Fei. Imagenet: A large-scale hierarchical image database. In *IEEE Conference on Computer Vision and Pattern Recognition*, pages 248–255. IEEE, 2009. 12

[9] Terrance DeVries and Graham W Taylor. Improved regularization of convolutional neural networks with cutout. *arXiv preprint arXiv:1708.04552*, 2017. 2

[10] James Diffenderfer, Brian R. Bartoldson, Shreya Chaganti, Jize Zhang, and Bhavya Kailkhura. A winning hand: Compressing deep networks can improve out-of-distribution robustness. In Marc'Aurelio Ranzato, Alina Beygelzimer, Yann N. Dauphin, Percy Liang, and Jennifer Wortman Vaughan, editors, *Advances in Neural Information Processing Systems 34: Annual Conference on Neural Information Processing Systems 2021, NeurIPS 2021, December 6-14, 2021, virtual*, pages 664–676, 2021. 2, 6

[11] N. Benjamin Erichson, Soon Hoe Lim, Francisco Utrera, Winnie Xu, Ziang Cao, and Michael W. Mahoney. Noisymix: Boosting robustness by combining data augmentations, stability training, and noise injections. *CoRR*, abs/2202.01263, 2022. 2, 7

[12] Robert Geirhos, Patricia Rubisch, Claudio Michaelis, Matthias Bethge, Felix A. Wichmann, and Wieland Brendel. Imagenet-trained cnns are biased towards texture; increasing shape bias improves accuracy and robustness. In *7th International Conference on Learning Representations, ICLR 2019, New Orleans, LA, USA, May 6-9, 2019*. OpenReview.net, 2019. 2, 7

[13] Tejas Gokhale, Rushil Anirudh, Bhavya Kailkhura, Jayaraman J Thiagarajan, Chitta Baral, and Yezhou Yang. Attribute-guided adversarial training for robustness to natural perturbations. In *Proceedings of the AAAI Conference on Artificial Intelligence*, volume 35, pages 7574–7582, 2021. 15

[14] Tejas Gokhale, Rushil Anirudh, Jayaraman J Thiagarajan, Bhavya Kailkhura, Chitta Baral, and Yezhou Yang. Improving diversity with adversarially learned transformations for domain generalization. In *Proceedings of the IEEE/CVF Winter Conference on Applications of Computer Vision*, pages 434–443, 2023. 2

[15] Kaiming He, Xiangyu Zhang, Shaoqing Ren, and Jian Sun. Deep residual learning for image recognition. In *Proceedings of the IEEE conference on computer vision and pattern recognition*, pages 770–778, 2016. 13

[16] Dan Hendrycks, Steven Basart, Norman Mu, Saurav Kadavath, Frank Wang, Evan Dorundo, Rahul Desai, Tyler Zhu, Samyak Parajuli, Mike Guo, Dawn Song, Jacob Steinhardt, and Justin Gilmer. The many faces of robustness: A critical analysis of out-of-distribution generalization. In *2021 IEEE/CVF International Conference on Computer Vision, ICCV 2021, Montreal, QC, Canada, October 10-17, 2021*, pages 8320–8329. IEEE, 2021. 7

[17] Dan Hendrycks and Thomas G. Dietterich. Benchmarking neural network robustness to common corruptions and perturbations. In *7th International Conference on Learning Representations, ICLR 2019, New Orleans, LA, USA, May 6-9, 2019*. OpenReview.net, 2019. 1, 2, 6, 12

[18] Dan Hendrycks, Norman Mu, Ekin Dogus Cubuk, Barret Zoph, Justin Gilmer, and Balaji Lakshminarayanan. Augmix: A simple data processing method to improve robustness and uncertainty. In *8th International Conference on Learning Representations, ICLR 2020, Addis Ababa, Ethiopia, April 26-30, 2020*. OpenReview.net, 2020. 2, 6, 7

[19] J. D. Hunter. Matplotlib: A 2d graphics environment. *Computing in Science & Engineering*, 9(3):90–95, 2007. 13

[20] Klim Kireev, Maksym Andriushchenko, and Nicolas Flammarion. On the effectiveness of adversarial training against common corruptions. In James Cussens and Kun Zhang, editors, *Uncertainty in Artificial Intelligence, Proceedings of the Thirty-Eighth Conference on Uncertainty in Artificial Intelligence, UAI 2022, 1-5 August 2022, Eindhoven, The Netherlands*, volume 180 of *Proceedings of Machine Learning Research*, pages 1012–1021. PMLR, 2022. 2, 6

[21] Thomas Kluyver, Benjamin Ragan-Kelley, Fernando Pérez, Brian Granger, Matthias Bussonnier, Jonathan Frederic, Kyle Kelley, Jessica Hamrick, Jason Grout, Sylvain Corlay, Paul Ivanov, Damián Avila, Safia Abdalla, Carol Willing, and Jupyter development team. Jupyter notebooks - a publishing format for reproducible computational workflows. In Fernando Loizides and Birgit Scmidt, editors, *Positioning and Power in Academic Publishing: Players, Agents and Agendas*, pages 87–90, Netherlands, 2016. IOS Press. 13

[22] Alex Krizhevsky. Learning multiple layers of features from tiny images. Technical report, University of Toronto, 2009. 12

[23] Alex Krizhevsky, Geoffrey Hinton, et al. Learning multiple layers of features from tiny images. 2009. 6

[24] Da Li, Yongxin Yang, Yi-Zhe Song, and Timothy M Hospedales. Deeper, broader and artier domain generalization. In *Proceedings of the IEEE international conference on computer vision*, pages 5542–5550, 2017. 2

[25] Apostolos Modas, Rahul Rade, Guillermo Ortiz-Jiménez, Seyed-Mohsen Moosavi-Dezfooli, and Pascal Frossard. PRIME: A few primitives can boost robustness to common corruptions. In Shai Avidan, Gabriel J. Brostow, Moustapha Cissé, Giovanni Maria Farinella, and Tal Hassner, editors, *Computer Vision - ECCV 2022 - 17th European Conference, Tel Aviv, Israel, October 23-27, 2022, Proceedings, Part XXV*, volume 13685 of *Lecture Notes in Computer Science*, pages 623–640. Springer, 2022. 2, 6

[26] Sajad Mousavi, Ricardo Luna Gutiérrez, Desik Rengarajan, Vineet Gundecha, Ashwin Ramesh Babu, Avisek Naug, Antonio Guillen, and Soumyendu Sarkar. N-critics: Self-refinement of large language models with ensemble of critics, 2023. 2

[27] Norman Mu and Justin Gilmer. Mnist-c: A robustness benchmark for computer vision. *arXiv preprint arXiv:1906.02337*, 2019. 2

[28] Arvind Neelakantan, Luke Vilnis, Quoc V Le, Ilya Sutskever, Lukasz Kaiser, Karol Kurach, and James Martens. Adding gradient noise improves learning for very deep networks. *arXiv preprint arXiv:1511.06807*, 2015. 9

[29] The pandas development team. pandas-dev/pandas: Pandas, Feb. 2020. 13

[30] Adam Paszke, Sam Gross, Francisco Massa, Adam Lerer, James Bradbury, Gregory Chanan, Trevor Killeen, Zeming Lin, Natalia Gimelshein, Luca Antiga, Alban Desmaison, Andreas Kopf, Edward Yang, Zachary DeVito, Martin Raison, Alykhan Tejani, Sasank Chilamkurthy, Benoit Steiner, Lu Fang, Junjie Bai, and Soumith Chintala. PyTorch: An Imperative Style, High-Performance Deep Learning Library. In H. Wallach, H. Larochelle, A. Beygelzimer, F. d'Alché-Buc, E. Fox, and R. Garnett, editors, *Advances in Neural Information Processing Systems 32*, pages 8024–8035. Curran Associates, Inc., 2019. 13

[31] Ben Poole, Jascha Sohl-Dickstein, and Surya Ganguli. Analyzing noise in autoencoders and deep networks. *arXiv preprint arXiv:1406.1831*, 2014. 9

[32] Raghavendran Ramakrishnan, Bhadrinath Nagabandi, Jose Eusebio, Shayok Chakraborty, Hemanth Venkateswara, and Sethuraman Panchanathan. Deep hashing network for unsupervised domain adaptation. In *Domain Adaptation in Computer Vision with Deep Learning*, pages 57–74. Springer, 2020. 2

[33] Soumyendu Sarkar, Ashwin Ramesh Babu, Vineet Gundecha, Antonio Guillen, Sajad Mousavi, Ricardo Luna, Sahand Ghorbanpour, and Avisek Naug. Rl-cam: Visual explanations for convolutional networks using reinforcement learning. In *Proceedings of the IEEE/CVF Conference on Computer Vision and Pattern Recognition*, pages 3860–3868, 2023. 2

[34] Soumyendu Sarkar, Ashwin Ramesh Babu, Vineet Gundecha, Antonio Guillen, Sajad Mousavi, Ricardo Luna, Sahand Ghorbanpour, and Avisek Naug. Robustness with query-efficient adversarial attack using reinforcement learning. In *Proceedings of the IEEE/CVF Conference on Computer Vision and Pattern Recognition*, pages 2329–2336, 2023. 2

[35] Soumyendu Sarkar, Ashwin Ramesh Babu, Sajad Mousavi, Sahand Ghorbanpour, Vineet Gundecha, Ricardo Luna Gutierrez, Antonio Guillen, and Avisek Naug. Reinforcement learning based black-box adversarial attack for robustness improvement. In *2023 IEEE 19th International Conference on Automation Science and Engineering (CASE)*, pages 1–8. IEEE, 2023. 2

[36] Soumyendu Sarkar, Ashwin Ramesh Babu, Sajad Mousavi, Vineet Gundecha, Sahand Ghorbanpour, Alexander Shmakov, Ricardo Luna Gutierrez, Antonio Guillen, and Avisek Naug. Robustness with black-box adversarial attack using reinforcement learning. In *AAAI 2023: Proceedings of the Workshop on Artificial Intelligence Safety 2023 (SafeAI 2023)*, volume 3381. https://ceur-ws.org/Vol-3381/8.pdf, 2023. 2

[37] Soumyendu Sarkar, Vineet Gundecha, Sahand Ghorbanpour, Alexander Shmakov, Ashwin Ramesh Babu, Alexandre Pichard, and Mathieu Cocho. Skip training for multi-agent reinforcement learning controller for industrial wave energy converters. In *2022 IEEE 18th International Conference on Automation Science and Engineering (CASE)*, pages 212–219. IEEE, 2022. 2

[38] Soumyendu Sarkar, Vineet Gundecha, Sahand Ghorbanpour, Alexander Shmakov, Ashwin Ramesh Babu, Avisek Naug, Alexandre Pichard, and Mathieu Cocho. Function approximation for reinforcement learning controller for energy from spread waves. In Edith Elkind, editor, *Proceedings of the Thirty-Second International Joint Conference on Artificial Intelligence, IJCAI-23*, pages 6201–6209. International Joint Conferences on Artificial Intelligence Organization, 8 2023. AI for Good. 2

[39] Soumyendu Sarkar, Vineet Gundecha, Alexander Shmakov, Sahand Ghorbanpour, Ashwin Ramesh Babu, Paolo Faraboschi, Mathieu Cocho, Alexandre Pichard, and Jonathan Fievez. Multi-objective reinforcement learning controller for multi-generator industrial wave energy converter. In *NeurIPs Tackling Climate Change with Machine Learning Workshop*, 2021. 2

[40] Soumyendu Sarkar, Vineet Gundecha, Alexander Shmakov, Sahand Ghorbanpour, Ashwin Ramesh Babu, Paolo Faraboschi, Mathieu Cocho, Alexandre Pichard, and Jonathan Fievez. Multi-agent reinforcement learning controller to maximize energy efficiency for multi-generator industrial wave energy converter. In *Proceedings of the AAAI Conference on Artificial Intelligence*, volume 36, pages 12135–12144, 2022. 2

[41] Soumyendu Sarkar, Sajad Mousavi, Ashwin Ramesh Babu, Vineet Gundecha, Sahand Ghorbanpour, and Alexander K Shmakov. Measuring robustness with black-box adversarial

attack using reinforcement learning. In *NeurIPS ML Safety Workshop*, 2022. 2

[42] Soumyendu Sarkar, Avisek Naug, Antonio Guillen, Ricardo Luna Gutierrez, Sahand Ghorbanpour, Sajad Mousavi, Ashwin Ramesh Babu, and Vineet Gundecha. Concurrent carbon footprint reduction (c2fr) reinforcement learning approach for sustainable data center digital twin. In *2023 IEEE 19th International Conference on Automation Science and Engineering (CASE)*, pages 1–8, 2023. 2

[43] Jiri Sedlar, Karla Stepanova, Radoslav Skoviera, Jan K Behrens, Matus Tuna, Gabriela Sejnova, Josef Sivic, and Robert Babuska. Imitrob: Imitation learning dataset for training and evaluating 6d object pose estimators. *IEEE Robotics and Automation Letters*, 8(5):2788–2795, 2023. 2

[44] Alexander Shmakov, Avisek Naug, Vineet Gundecha, Sahand Ghorbanpour, Ricardo Luna Gutierrez, Ashwin Ramesh Babu, Antonio Guillen, and Soumyendu Sarkar. Rtdk-bo: High dimensional bayesian optimization with reinforced transformer deep kernels. In *2023 IEEE 19th International Conference on Automation Science and Engineering (CASE)*, pages 1–8. IEEE, 2023. 2

[45] Karen Simonyan and Andrew Zisserman. Very deep convolutional networks for large-scale image recognition. *arXiv preprint arXiv:1409.1556*, 2014. 13

[46] Christian Szegedy, Wei Liu, Yangqing Jia, Pierre Sermanet, Scott Reed, Dragomir Anguelov, Dumitru Erhan, Vincent Vanhoucke, and Andrew Rabinovich. Going deeper with convolutions. In *Proceedings of the IEEE conference on computer vision and pattern recognition*, pages 1–9, 2015. 13

[47] Rui Tian, Zuxuan Wu, Qi Dai, Han Hu, and Yugang Jiang. Deeper insights into vits robustness towards common corruptions. *arXiv preprint arXiv:2204.12143*, 2022. 2, 7

[48] Guido Van Rossum and Fred L Drake Jr. *Python reference manual*. Centrum voor Wiskunde en Informatica Amsterdam, 1995. 13

[49] Vikas Verma, Alex Lamb, Christopher Beckham, Amir Najafi, Ioannis Mitliagkas, David Lopez-Paz, and Yoshua Bengio. Manifold mixup: Better representations by interpolating hidden states. In *International conference on machine learning*, pages 6438–6447. PMLR, 2019. 2

[50] Zekai Wang, Tianyu Pang, Chao Du, Min Lin, Weiwei Liu, and Shuicheng Yan. Better diffusion models further improve adversarial training. *CoRR*, abs/2302.04638, 2023. 2, 6

[51] Michael Lawrence Waskom. seaborn: statistical data visualization. *Journal of Open Source Software*, 60, April 2021. 13

[52] Zhenlin Xu, Deyi Liu, Junlin Yang, Colin Raffel, and Marc Niethammer. Robust and generalizable visual representation learning via random convolutions. *arXiv preprint arXiv:2007.13003*, 2020. 2

[53] Sangdoo Yun, Dongyoon Han, Seong Joon Oh, Sanghyuk Chun, Junsuk Choe, and Youngjoon Yoo. Cutmix: Regularization strategy to train strong classifiers with localizable features. In *Proceedings of the IEEE/CVF international conference on computer vision*, pages 6023–6032, 2019. 2

[54] Hongyi Zhang, Moustapha Cissé, Yann N. Dauphin, and David Lopez-Paz. mixup: Beyond empirical risk minimization. In *6th International Conference on Learning Representations, ICLR 2018, Vancouver, BC, Canada, April 30 - May 3, 2018, Conference Track Proceedings*. OpenReview.net, 2018. 2